\documentclass{IEEEoj}
\usepackage{booktabs}
\usepackage{multirow}
\usepackage{amsmath}
\usepackage{amssymb}
\usepackage{amsmath}
\usepackage{bm} 
\usepackage{color}
\usepackage[T1]{fontenc}
\usepackage{colortbl}

\usepackage{cite}
\usepackage{amsmath,amssymb,amsfonts}
\usepackage{algorithmic}
\usepackage{graphicx,color}
\usepackage{textcomp}
\def\BibTeX{{\rm B\kern-.05em{\sc i\kern-.025em b}\kern-.08em
    T\kern-.1667em\lower.7ex\hbox{E}\kern-.125emX}}
\AtBeginDocument{\definecolor{ojcolor}{cmyk}{0.93,0.59,0.15,0.02}}
\def\OJlogo{\vspace{-14pt}\includegraphics[height=28pt]{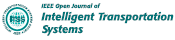}}

\begin{document}
\receiveddate{XX Month, XXXX}
\reviseddate{XX Month, XXXX}
\accepteddate{XX Month, XXXX}
\publisheddate{XX Month, XXXX}
\currentdate{XX Month, XXXX}
\doiinfo{TITS.2024.3375528}

\title{Visual-information-driven Model for Crowd Simulation Using Temporal Convolutional Network}

\author{Xuanwen Liang, Eric Wai Ming Lee \authorrefmark{1}}
\affil{Department of Architecture and Civil Engineering, City University of Hong Kong, Hong Kong}
\corresp{CORRESPONDING AUTHOR: Eric Wai Ming Lee (e-mail: ericlee@cityu.edu.hk).}
\authornote{The work described in this paper was fully supported by the grants from CityU Project No. 7005769 and 7005895.}
\markboth{Preparation of Papers for IEEE OPEN JOURNALS}{Author \textit{et al.}}

\begin{abstract}
Crowd simulations play a pivotal role in building design, influencing both user experience and public safety. While traditional knowledge-driven models have their merits, data-driven crowd simulation models promise to bring a new dimension of realism to these simulations. However, most of the existing data-driven models are designed for specific geometries, leading to poor adaptability and applicability. A promising strategy for enhancing the adaptability and realism of data-driven crowd simulation models is to incorporate visual information, including the scenario geometry and pedestrian locomotion. Consequently, this paper proposes a novel visual-information-driven (VID) crowd simulation model. The VID model predicts the pedestrian velocity at the next time step based on the prior social-visual information and motion data of an individual. A radar-geometry-locomotion method is established to extract the visual information of pedestrians. Moreover, a temporal convolutional network (TCN)-based deep learning model, named social-visual TCN, is developed for velocity prediction. The VID model is tested on three public pedestrian motion datasets with distinct geometries, i.e., corridor, corner and T-junction. Both qualitative and quantitative metrics are employed to evaluate the VID model, and the results highlight the improved adaptability of the model across all three geometric scenarios. Overall, the proposed method demonstrates effectiveness in enhancing the adaptability of data-driven crowd models.
\end{abstract}



\maketitle

\section{INTRODUCTION}
\IEEEPARstart{C}{rowd} simulations are widely used for modelling pedestrian movements, interactions and behaviours. They play a crucial role in facility design with the consideration of both user comfort and public safety. Crowd simulations allow building designers to understand the crowd movement and behaviour, enabling informed decision-making for enhancing the user-friendliness of facilities. Moreover, crowd simulations are of great importance in fire safety planning. In the event of a fire, the required safe egress time (RSET) must be less than the available safe egress time for occupants' safety. RSET values are typically determined through crowd simulations. These aspects highlight the importance of developing a realistic crowd simulation model.

\par
Various mature crowd simulation software tools have been used for industrial and research purposes. Most of these tools are based on knowledge-driven models, such as the social force (SF) model \cite{Helbing2000} and cellular automaton (CA) model \cite{VARAS2007631}. In such models, the pedestrian motion is ruled by physical formulas or expert knowledge. For instance, the SF model describes pedestrian movement subject to three types of forces: the self-driven force towards the exit, repulsive force from other pedestrians, and repulsive force from walls. By treating each pedestrian as a physical particle, their acceleration can be calculated using Newton's second law. The CA model divides the movement space into discrete grid cells, with the pedestrians moving to adjacent grid cells under the influence of static and dynamic fields. 

\par
Traditional knowledge-driven models, which have been extensively studied for over 20 years, have demonstrated notable success in macroscopic aspects. For example, the SF model can effectively reproduce the faster-is-slower phenomenon \cite{Helbing2000}. Additionally, both the SF model and CA model can reproduce typical pedestrian self-organization behaviors, e.g., lane formation in bidirectional flow \cite{10.1007/3-540-28091-X_36}, oscillatory flow \cite{Helbing_1995}, and arch formation \cite{Helbing2000} at bottlenecks. Despite their remarkable achievements, knowledge-driven models are often ineffective in the microscopic aspect, particularly in regards to the accuracy of pedestrian movement trajectories \cite{9043898}. This deficiency is attributable to the inherent limitations of the knowledge-driven fashion. More specifically, the mechanism of crowd movement is highly complex and affected by a wide range of factors, including but not limited to the interactions between pedestrians \cite{MA20102101, 10.1371/journal.pone.0010047}, individual personalities and psychology \cite{BELLOMO20161, WANG2023128411}, and the emergency level \cite{doi:10.1098/rsif.2016.0414}. However, the physical equations and expert experience employed in knowledge-driven models are overly simplistic and cannot adequately capture the complex mechanism of crowd movement.

\par
Data-driven crowd simulations represent a promising approach for addressing the aforementioned shortcomings. With advancements in image processing and machine learning technologies, the implementation of data-driven models has become increasingly feasible. Unlike knowledge-driven models, data-driven models are not governed by physical formulas or expert knowledge and instead learn pedestrian movement from large amounts of real-world motion data.
\par

Several researchers have explored the utility of data-driven crowd simulations and demonstrated their immense potential \cite{SONG2018827, ZHAO2020123825, 8790989, 9043898, 7464842}. 
Initial research was aimed at investigating simple single scenarios, such as pedestrian flow in corridors \cite{ZHAO2020123825} and pedestrians’ road crossing flow \cite{7464842}, using simple neural networks with dense layers. Various factors affecting the movement of the subject pedestrian, such as the velocity of the subject pedestrian and relative velocities of neighbouring pedestrians, were considered and used as the input features of the neural networks.  Following their work, \cite{ZHAO2021103260} proposed a multi-feature fusion recursive neural network (MFF-RNN) to simulate bidirectional flow in corridors. Superior performance was achieved using a more complex model structure. Furthermore, \cite{9898931} established a data-driven model for T-shape-geometry scenarios. 

\par
Despite the promising outcomes, the existing models exhibit several limitations. First, they are designed for specific single geometries, and thus their generalizability to other geometries is uncertain. Second, in many of the studies, pedestrian motion data from the testing scenarios were used to form the training dataset, resulting in the diminished practicality of the model in real-world applications. 

\par
Several researchers have attempted to enhance the adaptability and practicality of data-driven crowd simulation models \cite{9043898, YAO2020173}. For example, \cite{9043898} utilized a convolutional long short-term memory (LSTM) network. In their work, the test scenarios were independent of the training scenarios, and the model was validated for two types of geometries, i.e., room and corridor. \cite{YAO2020173} proposed a residual neural-network-based method to improve scene adaptability. Nevertheless, the adaptability of current data-driven models to diverse geometries remains inadequate. Models validated across two different geometric settings are rare, and those spanning a wider range of geometries are even rarer.

\par
Although certain studies have demonstrated the superior accuracy of data-driven models in predicting pedestrian trajectories in specific scenarios, the adaptability of these models is far weaker than that of knowledge-driven models. Knowledge-driven models, such as the SF and CA models, are highly adaptable and can be applied to nearly any geometry. To promote the application of data-driven models and leverage the advantage of enhanced realism, their adaptability must be demonstrated across various geometries.

\par
The development of a highly adaptive data-driven crowd simulation model presents several challenges. First, it's hard to calibrate the model parameters in data-driven models as it is done in knowledge-driven models. Enhancing the adaptability and accuracy of data-driven models is heavily reliant on the effectiveness of the feature inputs. Second, as discussed, pedestrian motion is complex and influenced by numerous factors, including psychological, social, physical and environmental aspect. Accurately selecting the key influential factors as input is challenging. Most previous studies focused on specific geometries and considered influential factors for only those geometries. Although this strategy results in excellent performance for the studied geometries, it limits the generalizability of the models to other geometries.

To improve the adaptability of crowd simulation models, we propose the incorporation of visual information, which includes the scene geometry and pedestrian locomotion. This information has often been overlooked in prior research, which focused on social and local interactions. Pedestrians can inherently perceive the geometry of the scenario and their own locomotion through visual cues. This information allows them to identify the locations of exits and walls and determine their overall heading direction while navigating. Previous research has shown that models centred on vision alone can effectively replicate collective behaviour \cite{doi:10.1126/sciadv.aay0792}. 

Predicting pedestrians' velocity at next time step based on input features from preceding time steps, with the aim of achieving crowd simulation, can be identified as a sequence modelling problem. This problem can be effectively addressed through a neural network. Typical neural networks for sequence modeling include LSTM \cite{6795963}, gated recurrent units (GRU) \cite{cho-etal-2014-properties} with recurrent architectures, and temporal convolutional network (TCN) \cite{bai2018empirical} utilizing convolutional approaches. While recurrent networks have traditionally dominated in sequence modeling, recent studies have shown that convolution-based TCNs outperform canonical recurrent networks such as LSTMs across diverse tasks and datasets \cite{bai2018empirical, 9671488}. Compared to LSTM and GRU, TCN exhibits a lower memory requirement during training. LSTMs and GRUs consume substantial memory to store partial results for their multiple cell gates. Furthermore, unlike recurrent networks where predictions for future time steps are dependent on the completion of preceding steps, TCN enables parallel processing due to utilizing the same filter in each layer. Given these advantages of TCN, we employ TCN for predicting pedestrian velocity.

To summarize, the development of a highly adaptive data-driven model is of significance and possesses great challenges. Visual information may be the key for improving the adaptability of such models. Considering these aspects, this paper proposes a novel visual-information-driven (VID) model. The VID model comprises three modules: a data processing (DP) module, a velocity prediction (VP) module, and a rolling forecast (RF) module. Particularly, the VP module is essentially a neural network based on the TCN \cite{bai2018empirical}, named social-visual TCN (SVTCN). The objective of this model is to improve the adaptability of data-driven models in crowd simulations by effectively capturing visual information. The contributions of this paper can be summarized as follows:

    \begin{figure*}[h]
    \centering
    \includegraphics[width=0.7\textwidth]{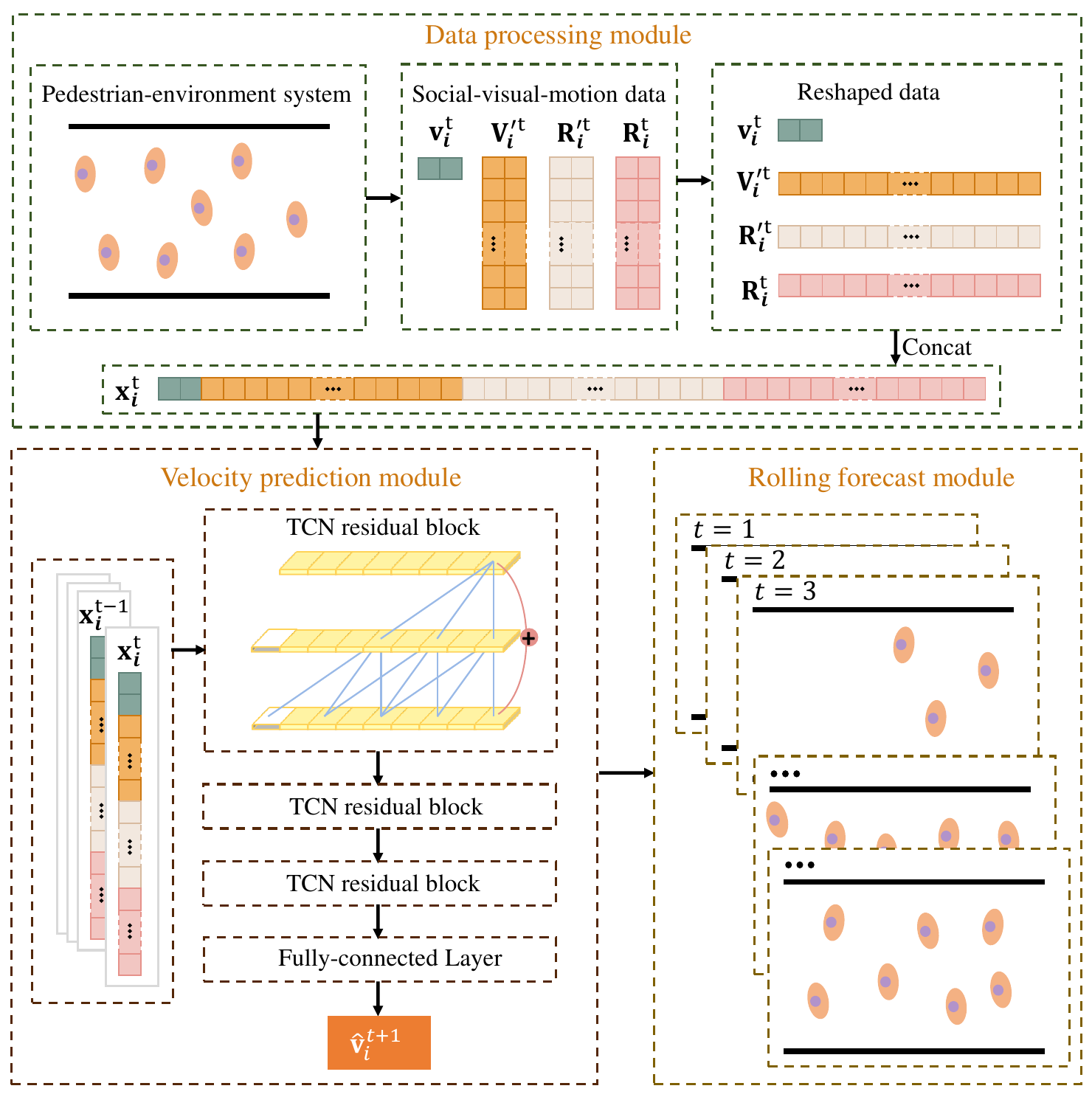}
    \caption{Overview of the visual-information-driven model. }
    \label{fig:vid_overview}
    \end{figure*}
    
\begin{itemize}
    \item We emphasize the importance of visual information, including the geometry of the scenario and pedestrian locomotion, in enhancing the adaptability and realism of data-driven crowd simulations.
    \item We propose a VID model for crowd simulations. In this model, we develop the radar-geometry-locomotion (RGL) method to extract visual information. Based on this, we establish a novel data-driven model named SVTCN.
    \item We conduct experiments on three publicly available datasets featuring distinct scenario geometries: corridor, corner and T-junction. The results demonstrate that the proposed VID model achieves satisfactory performance in terms of both qualitative and quantitative metrics.
\end{itemize}

\par
The remainder of the paper is structured as follows. Section 2 describes the proposed VID model. Section 3 details the experiments. Section 4 provides the simulation results. Section 5 discusses the findings. Section 6 presents the conclusions.

% In the early stage, \cite{7464842} simulated pedestrian motion in pedestrians' road crossing via simplest neural networks with linear layers. Although the studied scenario is really simple, it provides a pattern of input features of data-driven models. Specifically, the input features of their model include the velocity of the subject pedestrian, the relative velocity and positions of the neighbours, and  neighbouring pedestrians' relative velocities, as influential factors in the subject pedestrian's movement 
% 

\section{Model} \label{sec:model}

\subsection{Overview}
Considering we are given the historical trajectories of a group of pedestrians and environmental information, we aim to simulate the future motion of the crowd. To this end, the VID model takes the previous motion and environmental data as inputs, and it outputs the velocity of each pedestrian at the next time step. Fig. \ref{fig:vid_overview} presents an overview of the VID model, which consists of three modules: the DP, VP and RF module. The DP module is responsible for extracting pedestrian motion data and social-visual information. The extracted information is then fed to the VP module, which is a neural network known as SVTCN in this paper. The SVTCN learns the pedestrian motion dynamics from the input features and outputs the pedestrians’ velocities at the next time step. The network is trained by backpropagation algorithm \cite{werbos1975beyond}. After the completion of training, crowd simulation is realized using the trained network in the RF module.

\begin{figure*}[h]
\centering
\includegraphics[width=0.7\textwidth]{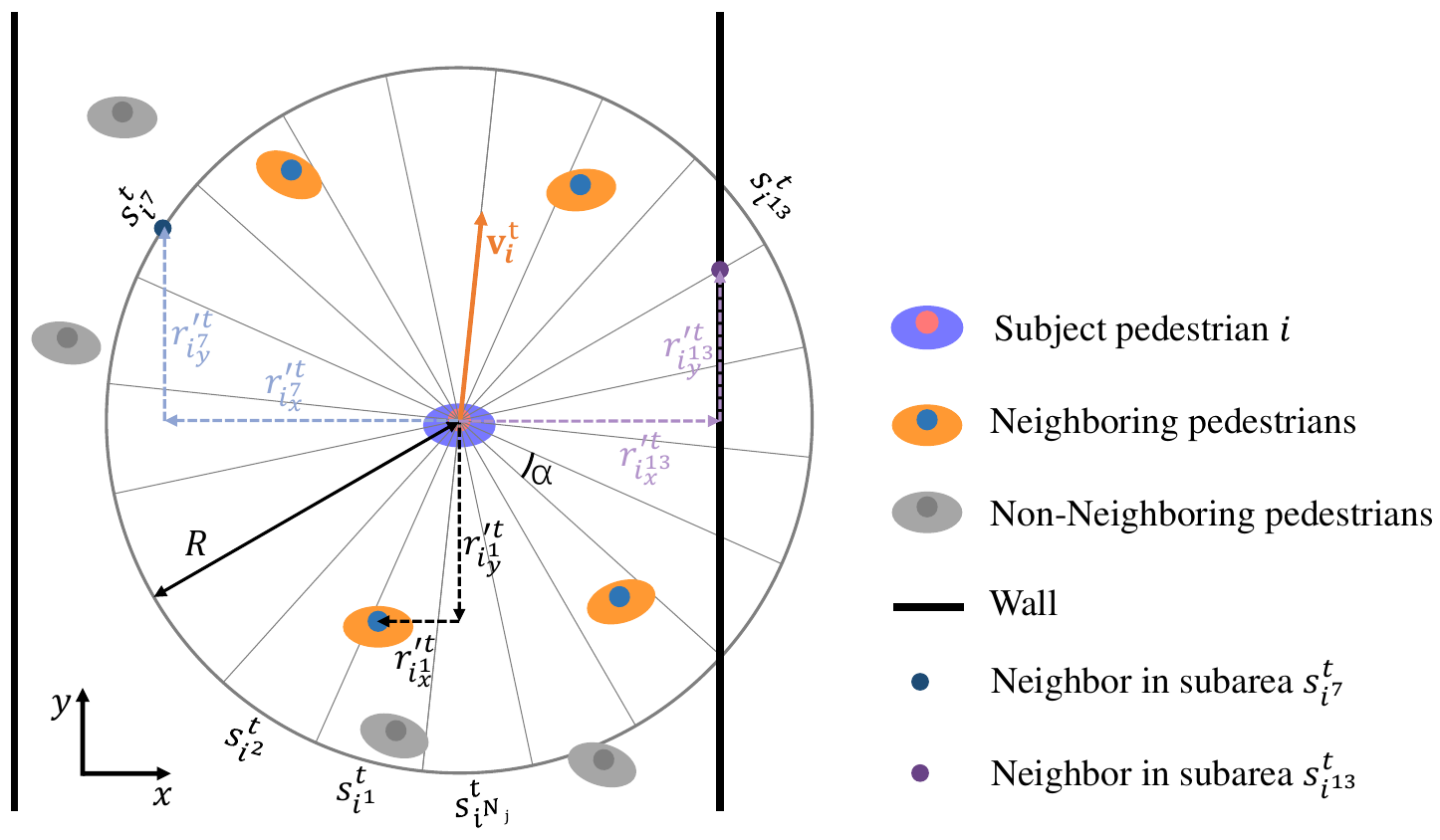}
\caption{Schematic of radar-nearest neighbours for subject pedestrian $i$ at time step $t$. }
\label{fig:schematic_rader_nn}
\end{figure*}

\subsection{Data processing (DP) module}\label{DP_module}
The DP module extracts the key features influencing pedestrian motion. Pedestrians and their surroundings constitute a pedestrian-environment system. The environmental stimuli affect the pedestrian motion and behaviour \cite{6251245}.  Although the environments may be diverse, some common stimuli affect pedestrian motion, stemming mainly from social and visual factors. Social stimuli include neighbouring pedestrians \cite{MA20102101}, resulting in following and avoiding behavior \cite{XIE2022105875}. Visual stimuli mainly come from walls and exits \cite{Helbing2000}. In addition to these environmental stimuli, pedestrians’ velocities are affected by their previous motion due to inertia effects \cite{Yanagisawa_2009}. Considering these aspects, the DP module extracts social-visual features and the pedestrians’ own motion from the pedestrian-environment system. Specifically, this set of information includes the velocity ${\mathbf{v}_i^t}$ of the subject pedestrian; relative positions ${\mathbf{R}^{'t}_i}$ and velocities $\mathbf{V}^{'t}_i$ of neighbours; and relative positions ${\mathbf{R}^{t}_i}$ of surrounding walls.

\begin{enumerate}
    \item Velocity of the subject pedestrian, ${\mathbf{v}_i^t}$. Empirical studies have highlighted that a pedestrian’s speed typically does not change abruptly \cite{Courtine2003}. A relaxation time \cite{LIU201744, MA20102160, Moussa_d_2009} is required for pedestrians to adapt their motion to the preferred speed. In other words, pedestrians gradually adjust their speeds to the desired value rather than making abrupt adjustments. Furthermore, with increases in pedestrians’ turning angles, their speeds tend to decrease \cite{Courtine2003}. Therefore, the changes in direction between the previous and subsequent states affect the subsequent speed. To summarize, a pedestrian’s future velocity is related to their past velocities, warranting their use as input features. The input feature ${\mathbf{v}_i^t} = [v_{i_x}^t, v_{i_y}^t] \in \mathbb{R}^{1 \times 2}$ is time-dependent, where $i$ represents the subject pedestrian. $v_{i_x}^t$ and $v_{i_y}^t$ are the magnitude components of the subject pedestrian's velocity in the $x$- and $y$-axis directions at time step $t$, respectively.

    \item Relative positions ${\mathbf{R}^{'t}_i}$ and velocities $\mathbf{V}^{'t}_i$ of neighbours. Pedestrians engage in social interactions with other pedestrians. For example, they tend to maintain suitable distances from neighbouring pedestrians to avoid collision \cite{RevModPhys.73.1067}, follow other pedestrians spontaneously \cite{XIE2022104100}, and exhibit competitive and cooperative behaviours \cite{drury2009cooperation}. Thus, the motion of the subject pedestrian is affected by other pedestrians, and vice versa. Considering all pedestrians in a given scenario may incur high computational costs. Furthermore, studies have suggested that by considering only the neighbouring pedestrians, the essential features and phenomena of pedestrian flow can be well reproduced \cite{PhysRevE.71.036121, MA20102101}. Common approaches to identify neighbouring pedestrians include the $k$-nearest-neighbour ($k$NN) method \cite{MA20102101, 8790989}  or interaction radius approach \cite{wkas2006social, PhysRevEYYF}. Specifically, the $k$NN method counts the $k$-nearest neighbours of the subject pedestrian, whereas the interaction radius approach focuses only on pedestrians within a certain metric distance, (i.e., the interaction radius). In this study, we use the radar-nearest-neighbour (Radar-NN) interaction pattern \cite{ZHAO2021103260}, which combines the interaction radius method and $k$NN, to determine neighbours. Subsequently, we extract the relative positions ${\mathbf{R}^{'t}_i}$ and velocities ${\mathbf{V}^{'t}_i}$ of the these neighbours as input features. This process is detailed in the following text.
    \par
    
    A schematic of the Radar-NN interaction pattern is shown in Fig. \ref{fig:schematic_rader_nn}. Let $R$ denote the social interaction radius. For each subject pedestrian $i$ at time step $t$ ($i, t$), a circular interaction area is determined. This area is divided into several equally-sized subareas with a unified central angle $\alpha$. Thus, the subareas for $(i, t)$ can be defined as ${\mathcal{S}_i^t} = \{s_{i^1}^t, s_{i^2}^t, \cdots, s_{i^j}^t, \cdots, s_{i^{N_j}}^t\}$, where $N_j=360^\circ/\alpha$ is the number of subareas, and $j$ represents the subarea index. The reverse extension of ${\mathbf{v}_i^t}$ is the boundary line between $s_{i^1}^t$ and $s_{i^{N_j}}^t$. For each subarea $s_{i^j}^t$, the subject pedestrian detects all other pedestrians or walls in $s_{i^j}^t$ at time step $t$ and considers the nearest entity (a pedestrian or wall) to them based on the Euclidean distance as neighbour $j$. In situations where there are no pedestrians or walls within $s_{i^j}^t$, the position of the neighbour $j$ is determined by the intersection of the perpendicular bisector of subarea $s_{i^j}^t$ and the interaction circle, as exemplified by the blue point for $s_{i^{7}}^t$ in Fig. \ref{fig:schematic_rader_nn}. Therefore, each subarea corresponds to one neighbour, and the number of subareas is equal to that of the radar-nearest neighbours. The index set of the radar-nearest neighbours for ($i, t$) is thus denoted as $\mathcal{J} = \{1, 2, \cdots, j, \cdots, N_j\}$, where each neighbour is indexed by $j$.
    
    \par
    After identifying all neighbours for ($i, t$), the relative positions ${\mathbf{R}^{'t}_i}$ and velocities ${\mathbf{V}^{'t}_i}$ of the neighbours are determined. In cases where the nearest entity is a wall, as exemplified by subarea $s_{i^{13}}^t$ in Fig. \ref{fig:schematic_rader_nn}, the position of the neighbour is determined by the nearest point on the wall within $s_{i^{13}}^t$ to the subject pedestrian $i$. Thus, the purple point in Fig. \ref{fig:schematic_rader_nn} represents the position of the neighbour in this case. We define the relative positions of neighbours, ${\mathbf{R}^{'t}_i}$, for ($i,t$) as ${\mathbf{R}^{'t}_i} = [\mathbf{r}^{'t}_{i^1}; \cdots; \mathbf{r}^{'t}_{i^j}; \cdots; \mathbf{r}^{'t}_{i^{N_j}}] \in \mathbb{R}^{{N_j} \times 2}$, where $\mathbf{r}^{'t}_{i^j} \in \mathbb{R}^{1 \times 2}$ represents the position of neighbour $j$ relative to pedestrian $i$. To be precise, $\mathbf{r}^{'t}_{i^j} = [r^{'t}_{i^j_x}, r^{'t}_{i^j_y}]$, where $r^{'t}_{i^j_x}$ and $r^{'t}_{i^j_y}$ represent the magnitude components of the relative position in the $x$- and $y$-axis directions at time step $t$, respectively. Similarly, we define ${\mathbf{V}^{'t}_i}$, for ($i,t$) as ${\mathbf{V}^{'t}_i} = [\mathbf{v}^{'t}_{i^1}; \cdots; \mathbf{v}^{'t}_{i^j}; \cdots; \mathbf{v}^{'t}_{i^{N_j}}] \in \mathbb{R}^{{N_j} \times 2}$, where $\mathbf{v}^{'t}_{i^j} = [v^{'t}_{i^j_x}, v^{'t}_{i^j_y}]$.

    \item Relative positions ${\mathbf{R}^{t}_i}$ of surrounding walls. Pedestrians in indoor environments can perceive the scenario geometry and their own locomotion through visual information. In this manner, pedestrians can accurately determine the relative positions of the walls and exits by optical flow. This information provides them with spatial constraints. Moreover, pedestrians can navigate and determine their global heading direction based on this information. Drawing inspiration from the Radar-NN method \cite{ZHAO2021103260}, we establish the RGL method to capture visual information.

    Fig. \ref{fig:RGL_rel_position} schematically illustrates the RGL method. In RGL, we obtain the positions of the surrounding walls relative to the subject pedestrian. Specifically, we define the forward motion zone as an infinite semicircle with the direction of pedestrian movement as its axis of symmetry. Walls in only the forward motion zone are considered. For each subject pedestrian $i$ at time step $t$ ($i, t$), the subject pedestrian is treated as the center, and they emit rays ${\mathcal{A}_i^t} = \{a_{i^1}^t, a_{i^2}^t, \cdots, a_{i^k}^t, \cdots, a_{i^{N_k}}^t\}$ at fixed intervals of angle $\beta$  within the forward motion zone. Here, $k$ represents the ray index, and ${N_k}=180^\circ/\beta + 1$ is the number of rays. Furthermore, $a_{i^1}^t$ and $a_{i^{N_k}}^t$ denote the pedestrian speed rotated 90° anticlockwise and clockwise, respectively. Then, we define an intersection points set ${\mathcal{O}_i^t} = \{o_{i^1}^t, o_{i^2}^t, \cdots, o_{i^k}^t, \cdots, o_{i^{N_k}}^t\}$, where $o_{i^k}^t$ represents the intersection point of emitting ray $a_{i^k}^t$ and the surrounding wall. Each emitting ray corresponds to an intersection point. If emitting ray $a_{i^k}^t$ does not intersect with any walls, it indicates that it points to the exit. In this case, the virtual intersection point $o_{i^k}^t$ is still located on ray $a_{i^k}^t$, and the distance between $o_{i^k}^t$ and the subject pedestrian is set to a virtual exit-distance parameter, denoted by a large number $D_e$. We define the relative positions of these intersection points, ${\mathbf{R}^{t}_i}$, for ($i,t$) as ${\mathbf{R}^{t}_i} = [\mathbf{r}^{t}_{i^1}; \cdots; \mathbf{r}^{t}_{i^k}; \cdots; \mathbf{r}^{t}_{i^{N_k}}] \in \mathbb{R}^{{N_k} \times 2}$, where $\mathbf{r}^{t}_{i^k} \in \mathbb{R}^{1 \times 2}$ represents the position of intersection point $o_{i^k}^t$ relative to pedestrian $i$. More precisely, $\mathbf{r}^{t}_{i^k} = [r^{t}_{i^k_x}, r^{t}_{i^k_y}]$, where $r^{t}_{i^k_x}$ and $r^{t}_{i^k_y}$ represent the magnitude components of the relative position in the $x$- and $y$-axis directions at time step $t$, respectively.

    \begin{figure}[h]
    \centering
    \includegraphics[width=0.42\textwidth]{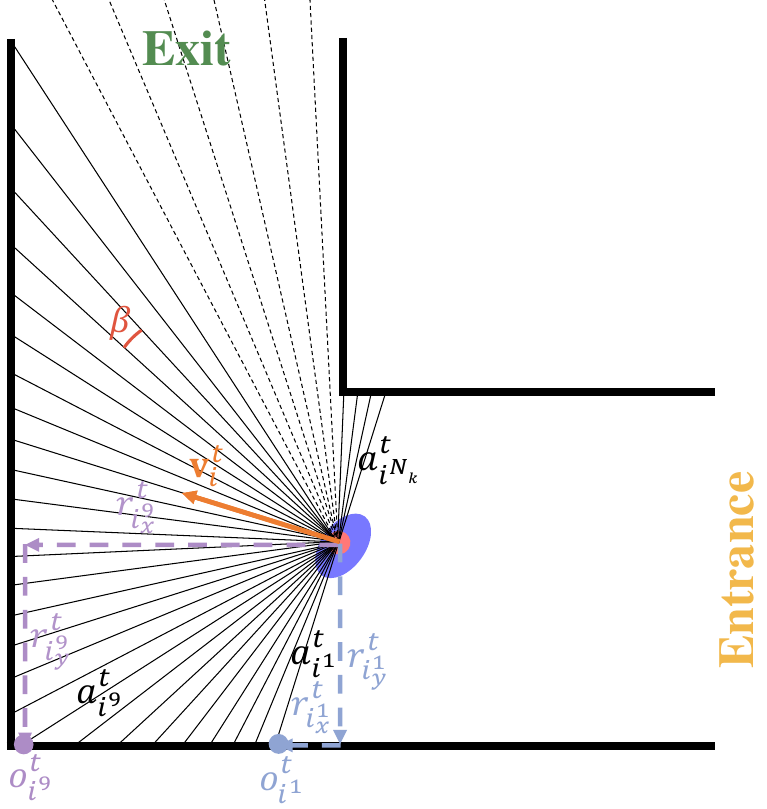}
    \caption{Schematic of the radar-geometry-locomotion method for subject pedestrian $i$ at time step $t$ in a corner scenario. $a_{i^k}^t$ is represented by solid and dotted black lines when it intersects with a wall and exit, respectively.}
    \label{fig:RGL_rel_position}
    \end{figure}

\end{enumerate} 
    \par
Lastly, we reshape ${\mathbf{v}_i^t}$, $\mathbf{V}^{'t}_i$, ${\mathbf{R}^{'t}_i}$ and ${\mathbf{R}_i^t}$ to one-dimensional (1D) vectors and concatenate them to vector $\mathbf{x}_i^t$, as shown in Fig. \ref{fig:vid_overview}.

%     Pedestrians' movement and navigation in indoor environment is greatly affected by the geometry of the architectures. On the one hand, their movement is subject to spatial constraints due to walls. They can't go through the walls. On the other hand, pedestrians' navigational strategies are strongly related to geometric cues. Most such strategies are so common that they do not occur to us when we perform them \cite{Werner2003CognitionML}. For instance, we rarely realize that the optical flows give us rich information about where we are headed when waiking down a hallway (Gibson, 1979).
% Tasks such as identifying a place or direction, retracing one’s path, or navigating a large-scale space, are essential elements to mobile organisms.cues for spatial orientation and wayfinding.  human spatial cognition.\par <Cognition Meets Le Corbusier - Cognitive Principles of Architectural Design >明天重点看。《Influence of Geometry and Objects on Local Route Choices during Wayfinding。明天重点看。》

\begin{figure*}[h]
\centering
\includegraphics[width=0.7\textwidth]{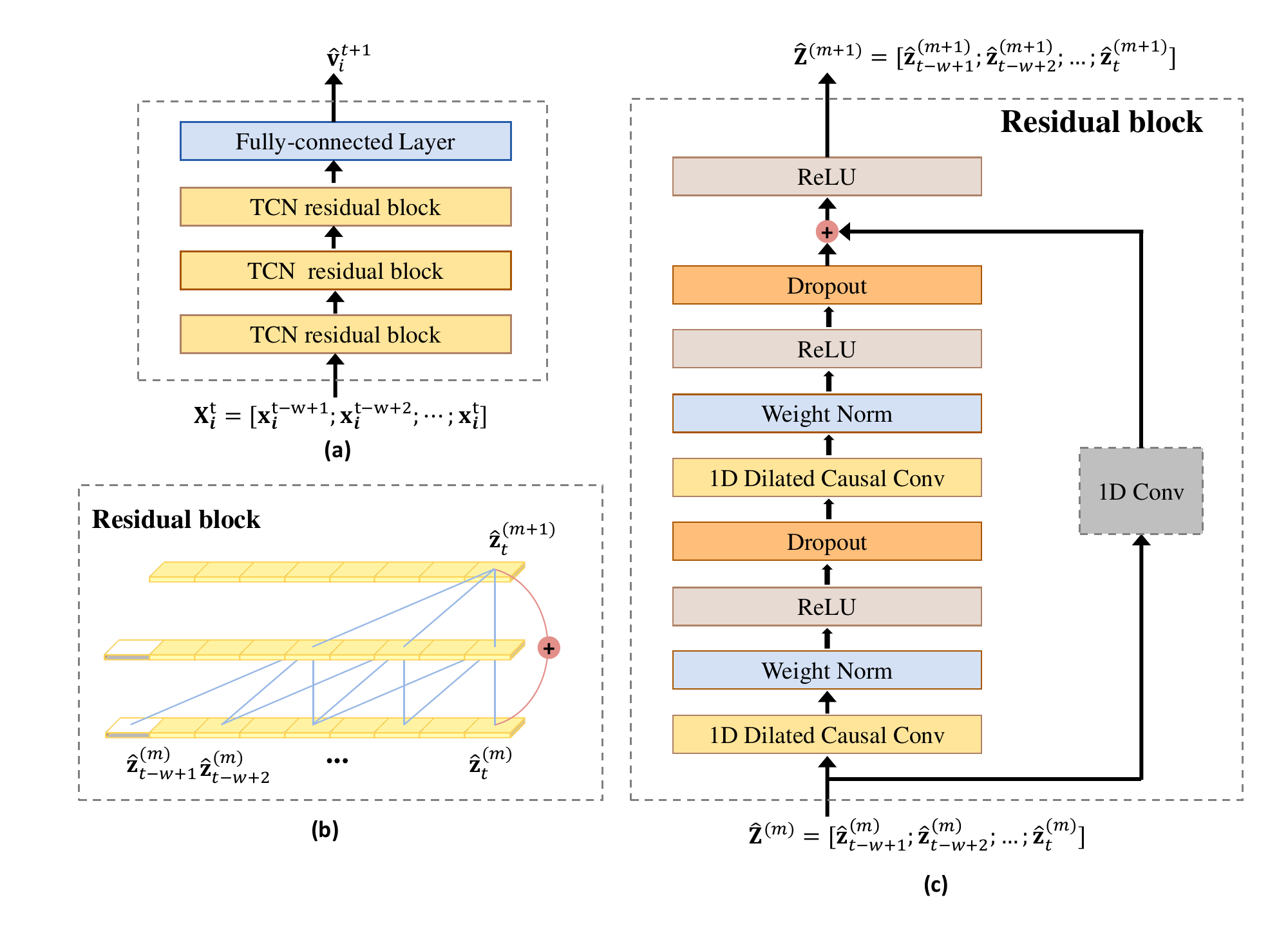}
\caption{Schematic of the VP module. (a) Structure of the VP module. (b) An example of residual connection in a TCN residual block with dilation factor $h = 2$ and filter size $q = 3$. (c) Structure of the TCN residual block proposed by \cite{bai2018empirical}.}
\label{fig:VP_module}
\end{figure*}

\subsection{Velocity prediction (VP) module}
The VP module predicts the subject pedestrian's velocity at the next time step based on their previous social-visual features and motion data extracted in the DP module. This task can be regarded as a sequence modelling problem and can be effectively solved by a  neural network. Here, we employ a TCN-based \cite{bai2018empirical} neural network to address it.

% Considering Temporal Convolutional Network (TCN) \cite{bai2018empirical}, which is designed for sequence modelling, has shown advanced performance in several tasks \cite{bai2018empirical}, we employ TCN in this work.
% Networks for sequence modelling encompass (Long Short-Term Memory) \cite{6795963}, GRU (Gated Recurrent Unit) \cite{cho-etal-2014-properties}, TCN (Temporal Convolutional Network) \cite{bai2018empirical}, etc. 
\par
The structure of the VP module, which is essentially a neural network (i.e., SVTCN), is shown in Fig. \ref{fig:VP_module} (a). The SVTCN consists of three TCN residual blocks \cite{bai2018empirical} and one fully-connected layer. It takes $\mathbf{X}_i^t = [\mathbf{x}_i^{t-w+1}; \mathbf{x}_i^{t-w+2}; \cdots; \mathbf{x}_i^{t}]$ as the input and outputs the predicted velocity of the subject pedestrian at the next time step, ${\hat{\mathbf{v}}_i^{t+1}} = [\hat{v}_{i_x}^{t+1}, \hat{v}_{i_y}^{t+1}] \in \mathbb{R}^{1 \times 2}$. Here, $w$ represents the length of the lookback window. 
\par

The key component of the VP module is the TCN residual block, as illustrated in Fig. \ref{fig:VP_module}. The block is constructed based on two fundamental principles. First, the input and output sequences of the TCN block are of equal length. Second, the block predicts the output for any time step using only the inputs that have been observed before. To fulfil the first criterion, the TCN block introduces a 1D convolutional network, and zero-padding is applied to ensure consistency in the input and output lengths across each layer. To fulfil the second criterion, the TCN block uses causal convolutions, which ensure that the output at time step $t$ in the subsequent layer is convolved only with the elements at or preceding time step $t$ in the previous layer. In addition, dilated convolution is applied to leverage the long-term memory capacity of the TCN. In summary, the TCN uses 1D dilated causal convolutions, which serves as the critical component of the TCN block. The 1D dilated causal convolution operation $\mathbf{\Theta}$ on element $e$ in the sequence is defined as
\begin{equation}
    \mathbf{\Theta}(e) = \sum_{g=0}^{q-1}f(g)\cdot \bm{Z}_{e-h \cdot g}
\end{equation}
where $f$ is the convolution kernel, $q$ is the kernel size, $h$ is the dilation factor, and $\bm{Z}$ is the input sequence of the convolutional operation. In this dilated convolution, the convolutional layer has a historical memory with length $h \cdot (q-1) +1$. When $h = 2^{m-1}$, where $m$ refers to the $m^{th}$ TCN residual block, the historical memory length increases exponentially with the increase in the network depth.

\par
The structure of each TCN residual block is shown in Fig. \ref{fig:VP_module} (c). Let $\hat{\mathbf{Z}}^{(m)}$ denote the input of the $m^{th}$ residual block. Then, the output of the block  $\hat{\mathbf{Z}}^{(m+1)}$ is calculated as
\begin{equation}
    \hat{\mathbf{Z}}^{(m+1)} = \text{ReLU}(\Phi(\hat{\mathbf{Z}}^{(m)}) + \Psi(\hat{\mathbf{Z}}^{(m}))
\end{equation}
where $\Phi$ and $\Psi$ are transformation operations. Specifically, $\Phi$ consists of a series of operations, including two rounds of 1D dilated causal convolutions, weight normalization \cite{salimans2016weight}, rectified linear unit (ReLU) activation \cite{nair2010rectified}, and dropout \cite{srivastava2014dropout}. When the input and output sequences have differing channel dimensions, $\Psi$ represents a 1D convolution that ensures consistency in the channel dimensions. When the channel dimensions are identical, $\Psi(\hat{\mathbf{Z}}^{(m)}) = \hat{\mathbf{Z}}^{(m)}$.

\par
Overall, the VP module outputs the predicted velocity of the subject pedestrian at the next time step, denoted as ${\hat{\mathbf{v}}_i^{t+1}}$. Let $\mathbf{v}_i^{t+1}$ denote the ground truth. By minimizing the sum of $\|{\hat{\mathbf{v}}_i^{t+1}} - \mathbf{v}_i^{t+1} \|$ for all training samples using the gradient descent algorithm, the network parameters can be trained.

    \begin{figure}[htpb]
    \centering
    \includegraphics[width=0.42\textwidth]{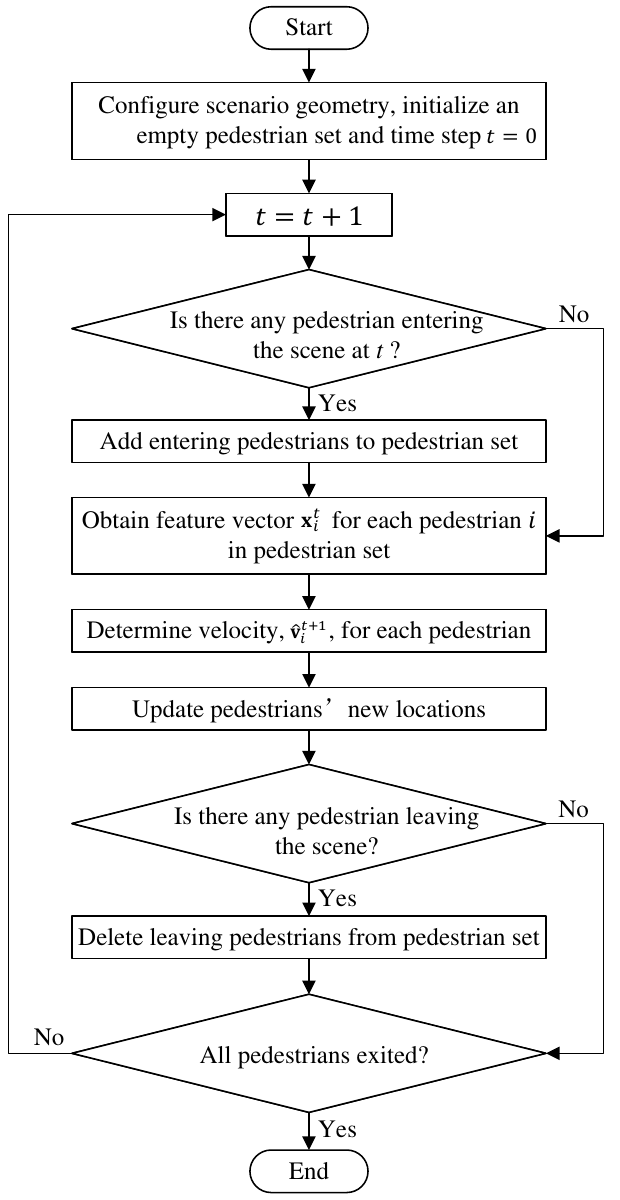}
    \caption{Flow chart of the RF procedure.}
    \label{fig:RF_floe_chart}
    \end{figure}

    \begin{figure*}[htpb]
    \centering
    \includegraphics[width=0.7\textwidth]{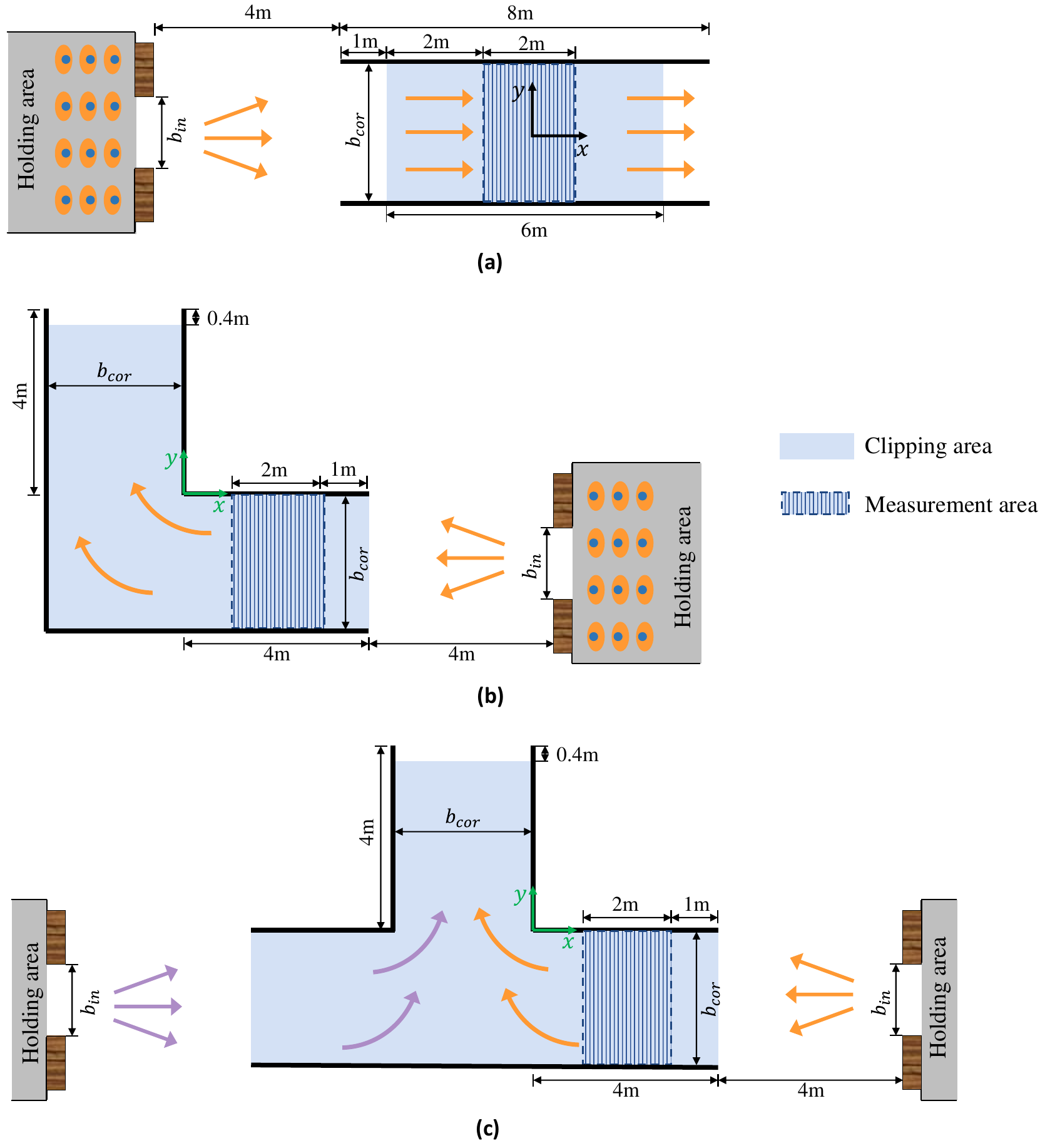}
    \caption{Schematics of the experimental setups (refer to https://ped.fz-juelich.de/da/doku.php). (a) Corridor. (b) Corner. (c) T-junction.}
    \label{fig:exp_setup}
    \end{figure*}

\subsection{Rolling forecast (RF) module}
% The RF module aims to simulate continues crowd movement. Once the network is trained, we can utilize the trained network to simulate crowd motion. Particularly, put pedestrian into the prediction scene, and keep the first 

The RF module simulates the continuous crowd movement. Once the network has been trained, it can be used to simulate pedestrian movements in simulation scenarios. The simulation scenarios are designed based on controlled experiments, i.e., the testing dataset in Tables \ref{tab:corridor_dataset}-\ref{tab:T-junction_dataset}, to accurately evaluate the simulation performance in terms of realism. Specifically, the geometries of the simulation scenarios, number of pedestrians, entry time of each pedestrian, and pedestrian positions during the initial $w$ (lookback window) time steps in the simulation scenarios are identical to those in the corresponding controlled experiments. Detailed information regarding the setup of the simulation scenarios can be found in Section \ref{sec:experiments}. 

\par
Fig. \ref{fig:RF_floe_chart} illustrates the flow chart of the RF process. Initially, the simulation scenario geometry is configured, and an empty set of pedestrians, $\mathcal{P}$, is created to represent the pedestrians within the scenario. Then, for time step $t$, if there are pedestrians entering the scenario at $t$, they are added to set $\mathcal{P}$. Next, the feature vector ${\mathbf{x}_i^t}$ is obtained for each pedestrian $i$ within set $\mathcal{P}$ at time step $t$, following the procedure described in the DP module. The simulated pedestrian velocity ${\hat{\mathbf{v}}_i^{t+1}}$ is then determined. Specifically, for each pedestrian $i$ in set $\mathcal{P}$, if the number of time steps since pedestrian $i$ entered the scene is equal to or greater than the length of the lookback window $w$, ${\hat{\mathbf{v}}_i^{t+1}}$ is predicted by the trained model using the feature sequence $\mathbf{X}_i^t = [\mathbf{x}_i^{t-w+1}; \mathbf{x}_i^{t-w+2}; \cdots; \mathbf{x}_i^{t}]$. Otherwise, ${\hat{\mathbf{v}}_i^{t+1}}$ is set equal to the velocity ${\mathbf{v}}_i^{t+1}$ observed in the controlled experiment. Subsequently, the locations of all pedestrians in set $\mathcal{P}$ are updated, and any pedestrians leaving the scenario are removed from set $\mathcal{P}$. This iterative procedure is repeated until all pedestrians have exited the scenario.

Similar to previous studies \cite{ZHAO2021103260, ZHAO2020123825, 9043898}, a quite small number of trajectories cross the scenario boundaries within our simulation process. This is primarily attributed to the absence of strict boundary constraints, as typically observed in the SF or CA models. We employ similar methods as presented in \cite{ZHAO2021103260, ZHAO2020123825}. Specifically, when a pedestrian traverses the scenario boundary, we reset their velocities for the preceding $w$-length time steps, guiding them to move forward while distancing themselves from the scenario boundary. Simultaneously, we recalculate the corresponding feature vectors for the preceding $w$-length time steps.

    \begin{figure*}[htpb]
    \centering
    \includegraphics[width=0.98\textwidth]{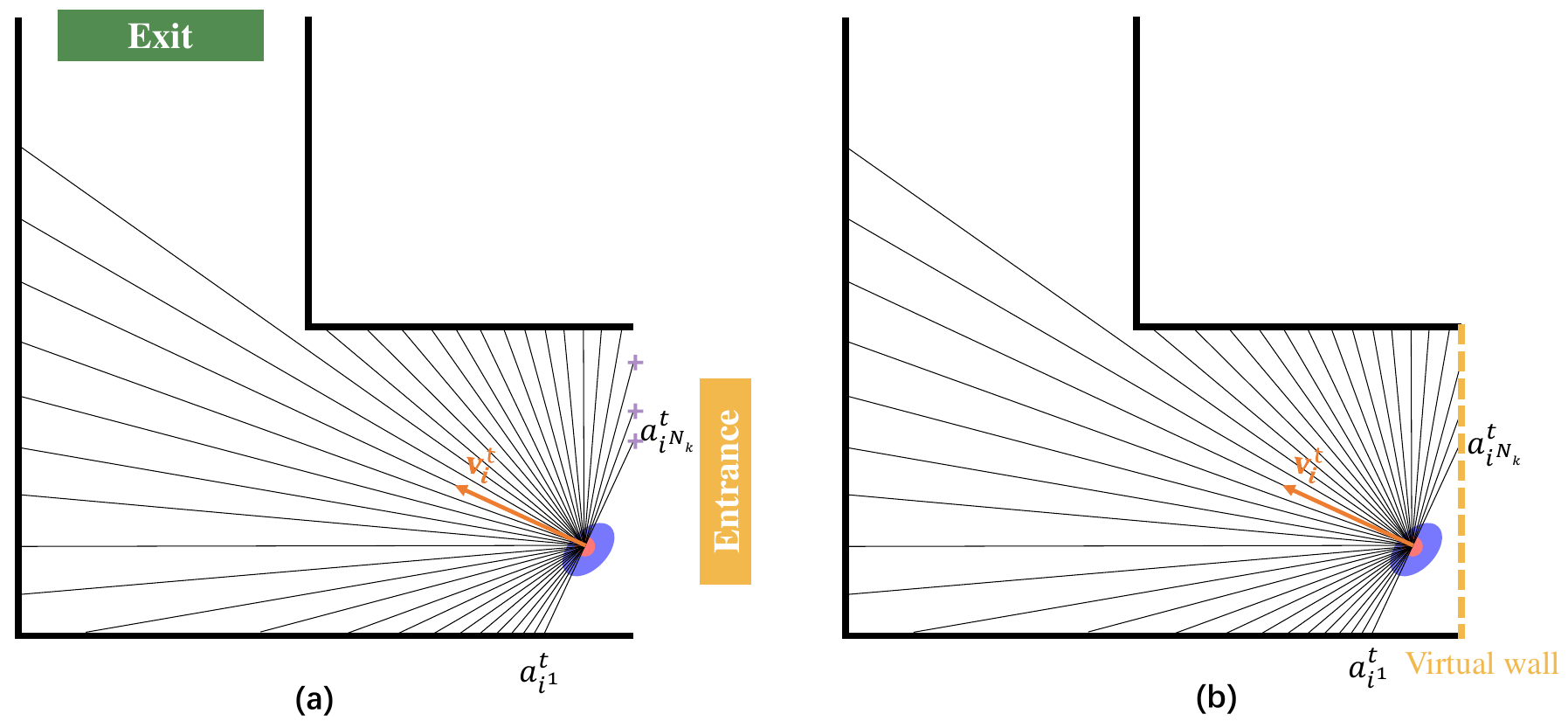}
    \caption{Entrance processing. (a) Intersections of emitted rays and the entrance, with the purple cross symbols denoting the respective intersection locations. (b) Implementation of a virtual wall within the entrance area for corner geometry. The yellow dotted line represents the virtual wall. }
    \label{fig:entrance_processing}
    \end{figure*}

\section{Experiments} \label{sec:experiments}
\subsection{Datasets} \label{sec:datasets}
Corridors, corners and T-junctions are among the most fundamental and prevalent elements in building design. They serve as essential components in complex architectural structures, encapsulating the majority of pedestrian flow patterns, including merging flow, turning flow, bottleneck flow and straightforward corridor flow. Given their ubiquity and the variety of pedestrian dynamics they represent, these three structures have been extensively studied in the field of pedestrian dynamics and evacuation research \cite{Cao_2017, su11195501, Ye_2019, Zhang_2011, LI2022126593}. Therefore, we select these three geometries to form our dataset. The Institute of Civil Safety Research at the Research Centre Jülich, Germany has conducted crowd movement experiments in these geometries under laboratory conditions. Details of these experiments, including the experimental setup, video recordings and pedestrian trajectory data, are available online (at https://ped.fz-juelich.de/da/doku.php). For our study, we utilize the trajectory data directly downloaded from this source to form our datasets, which are detailed in the following text.

\begin{enumerate}
\item Corridor dataset. Fig. \ref{fig:exp_setup}(a) illustrates the setup for the corridor experiment. This configuration includes a corridor with a length of 8m. In each run, pedestrians depart from the holding area, traverse through the corridor, and exit it. Multiple runs are conducted by adjusting the widths of the entrance ($b_{in}$) and corridor ($b_{cor}$) to regulate the crowd density. To enhance the quality of the dataset, only trajectory data for a clipping area with a length of 6m are used.

\item Corner dataset. Fig. \ref{fig:exp_setup}(b) illustrates the setup for the corner experiment. The widths of the entrance ($b_{in}$) and corridor ($b_{cor}$) are adjusted in each run to regulate the density in the corner. A clipping area is applied such that trajectories within a range of 0.4m near the exit are discarded.

\item T-junction dataset. As illustrated in Fig. \ref{fig:exp_setup}(c), pedestrians enter the T-junction from two sides and exit it from the top exit. The widths of the entrance ($b_{in}$) and corridor ($b_{cor}$) are varied in each run. A clipping area is also applied to ensure data quality.

\end{enumerate}

\subsection{Experimental settings}
\subsubsection{Entrance processing} \label{sec:entrance_processing}
Given that the emitted rays may intersect with the entrance, as shown in Fig. \ref{fig:entrance_processing}(a), the entrances are additionally processed during the extraction of social-visual information. Specifically, we introduce a virtual wall for each entrance. In the corridor and corner scenarios, a virtual wall is added to the entrance, as illustrated in Fig. \ref{fig:entrance_processing}(b). In the T-junction scenario, two virtual walls are added, one for each entrance. During the extraction of social-visual information, the virtual wall is considered equivalent to the physical wall.

\subsubsection{Dataset segmentation}
As described in Part \ref{sec:datasets} of Section \ref{sec:experiments}, multiple runs were conducted in the corridor, corner, and T-junction experiments by varying the widths of the entrance ($b_{in}$) and corridor ($b_{cor}$). For each of these geometries, the experiment scenarios are divided into two categories: training-validation scenarios and testing scenarios, as presented in Tables \ref{tab:corridor_dataset}-\ref{tab:T-junction_dataset}. Notably, the training-validation scenarios and testing scenarios were completely independent to demonstrate the generalizability of the proposed model. Furthermore, the parameters related to the geometries, specifically the widths $b_{cor}$ and $b_{in}$, have a significant impact on the density and pedestrian flow rate within the scenarios. When segmenting the datasets, we intentionally chose different widths for the training-validation set and the testing set scenarios to highlight the adaptability of the model. To illustrate, the widths of the corridors in the training-validation dataset are 1.8m and 2.4m, while in the testing dataset, it is 3m. Similarly, the width of the corners and T-junctions in the training-validation dataset is 2.4m, and in the testing dataset, it is 3m. Moreover, both the training-validation dataset and testing dataset encompass a wide range of pedestrian densities, ranging from small to large, achieved by varying $b_{in}$ and $b_{cor}$. This meticulous dataset selection process serves to further demonstrate the robustness of the model.

Using the method described in Section \ref{sec:model} and the entrance processing method, we extract input and target vectors from the training-validation scenarios to establish the training-validation set for each geometry. Prior to extraction, the trajectories within the corridors undergo a smoothing treatment. Taking inspiration from \cite{Cao_2018}, we employ the Savitzky-Golay filter \cite{Savitzky_smooth} for trajectory smoothing. Subsequently, these three training-validation sets are partitioned into training and validation sets with a ratio of 4:1, respectively. The training set is used to update the network parameters, whereas the validation set is used to identify the optimal model. The velocities specified in Section \ref{sec:model} are obtained from the trajectories. Specifically, the velocity components for pedestrian $i$ at time step $t$ in the $x$-axis, $v_{i_x}^t$, and $y$-axis, $v_{i_y}^t$ , are calculated as
\begin{equation}
v_{i_x}^t = \frac{x_i^t - x_i^{t-\Delta t}}{\Delta t}, \quad v_{i_y}^t = \frac{y_i^t - y_i^{t-\Delta t}}{\Delta t}
\end{equation}
Here, $x_i^t$ and $y_i^t$ represent the positions of pedestrian $i$ at time step $t$ along the $x$- and $y$-axes, respectively; and $\Delta t = 0.5s$ is the time interval. The frame rate for all experimental video recordings is 16 frames per second.

\begin{table*}[htb]
    \centering
    \footnotesize
    \caption{Parameters of the corridor dataset.}
    \begin{tabular}{
    >{\centering}m{0.2\textwidth} 
    >{\centering}m{0.25\textwidth} >{\centering}m{0.1\textwidth} 
    >{\centering}m{0.1\textwidth} >{\centering\arraybackslash}m{0.1\textwidth}}
      \toprule %Add a bold line to the header
      Dataset  & Name & $b_{in}[m]$ & $b_{cor}[m]$ & N$[Pers.]$ \\ 
      \hline
     \multirow{6}{*}{Training-validation} & 
                           corr-050-180 & 0.50 & 1.80 & 61 \\
                         \cline{2-5}
                          & corr-060-180 & 0.60 & 1.80 & 66 \\
                         \cline{2-5}
                          & corr-070-180 & 0.70 & 1.80 & 111 \\
                         \cline{2-5}
                          & corr-100-180 & 1.00 & 1.80 & 121 \\
                         \cline{2-5}
                          & corr-145-180 & 1.45 & 1.80 & 175 \\
                         \cline{2-5}
                          & corr-180-180 & 1.80 & 1.80 & 220 \\
                         \cline{2-5}
                         & corr-065-240 & 0.65 & 2.40 & 70 \\
                         \cline{2-5}
                          & corr-080-240 & 0.80 & 2.40 & 118 \\
                         \cline{2-5}
                          & corr-095-240 & 0.95 & 2.40 & 108 \\
                         \cline{2-5}
                          & corr-145-240 & 1.45 & 2.40 & 155 \\
                         \cline{2-5}
                         & corr-190-240 & 1.90 & 2.40 & 218 \\
                         \cline{2-5}
                          & corr-240-240 & 2.40 & 2.40 & 246 \\
    \hline
         \multirow{6}{*}{Testing} & 
                          corr-080-300 & 0.80 & 3.00 & 119 \\
                         \cline{2-5}
                          & corr-100-300 & 1.00 & 3.00 & 100 \\
                         \cline{2-5}
                          & corr-120-300 & 1.20 & 3.00 & 163 \\
                         \cline{2-5}
                          & corr-180-300 & 1.80 & 3.00 & 208 \\
                         \cline{2-5}
                          & corr-240-300 & 2.40 & 3.00 & 296 \\
                         \cline{2-5}
                          & corr-300-300 & 3.00 & 3.00 & 349 \\

    \bottomrule
    \end{tabular}
    \label{tab:corridor_dataset}
\end{table*}

\begin{table*}[htb]
    \centering
    \footnotesize
    \caption{Parameters of the corner dataset.}
    \begin{tabular}{
    >{\centering}m{0.2\textwidth} 
    >{\centering}m{0.25\textwidth} >{\centering}m{0.1\textwidth} 
    >{\centering}m{0.1\textwidth} >{\centering\arraybackslash}m{0.1\textwidth}}
      \toprule %Add a bold line to the header
      Dataset  & Name & $b_{in}[m]$ & $b_{cor}[m]$ & N$[Pers.]$ \\ 
      \hline
     \multirow{6}{*}{Training-validation} & 
                           corn-050-240 & 0.50 & 2.40 & 64 \\
                         \cline{2-5}
                          & corn-060-240 & 0.60 & 2.40 & 56 \\
                         \cline{2-5}
                          & corn-080-240 & 0.80 & 2.40 & 111 \\
                         \cline{2-5}
                          & corn-100-240 & 1.00 & 2.40 & 106 \\
                         \cline{2-5}
                          & corn-150-240 & 1.50 & 2.40 & 107 \\
                         \cline{2-5}
                         & corn-240-240 & 2.40 & 2.40 & 161 \\ 
    \hline
         \multirow{6}{*}{Testing} & 
                          corn-050-300 & 0.50 & 3.00 & 65 \\
                         \cline{2-5}
                          & corn-060-300 & 0.60 & 3.00 & 57 \\
                         \cline{2-5}
                          & corn-080-300 & 0.80 & 3.00 & 104 \\
                         \cline{2-5}
                          & corn-100-300 & 1.00 & 3.00 & 118 \\
                         \cline{2-5}
                          & corn-150-300 & 1.50 & 3.00 & 137 \\
                         \cline{2-5}
                          & corn-300-300 & 3.00 & 3.00 & 222 \\ 

    \bottomrule
    \end{tabular}
    \label{tab:corner_dataset}
\end{table*}

\begin{table*}[htb]
    \centering
    \footnotesize
    \caption{Parameters of the T-junction dataset.}
    \resizebox{0.9\textwidth}{!}{
    \begin{tabular}{
    >{\centering}m{0.2\textwidth}  
    >{\centering}m{0.25\textwidth} >{\centering}m{0.1\textwidth} 
    >{\centering}m{0.1\textwidth} >{\centering\arraybackslash}m{0.1\textwidth}}
      \toprule %Add a bold line to the header
      Dataset  & Name & $b_{in}[m]$ & $b_{cor}[m]$ & N$[Pers.]$ \\ 
      \hline
     \multirow{6}{*}{Training-validation} & 
                          T-junc-050-240 & 0.50 & 2.40 & 134 \\
                         \cline{2-5}
                          & T-junc-060-240 & 0.60 & 2.40 & 132 \\
                         \cline{2-5}
                          & T-junc-080-240 & 0.80 & 2.40 & 228 \\
                         \cline{2-5}
                          & T-junc-100-240 & 1.00 & 2.40 & 208 \\
                         \cline{2-5}
                          & T-junc-120-240 & 1.20 & 2.40 & 305 \\
                         \cline{2-5}
                          & T-junc-150-240 & 1.50 & 2.40 & 305 \\
                         \cline{2-5}
                          & T-junc-240-240 & 2.40 & 2.40 & 303 \\ 
    \hline
         \multirow{6}{*}{Testing} & 
                          T-junc-050-300 & 0.50 & 3.00 & 146 \\
                         \cline{2-5}
                          & T-junc-080-300 & 0.80 & 3.00 & 207 \\
                         \cline{2-5}
                          & T-junc-120-300 & 1.20 & 3.00 & 306 \\
                         \cline{2-5}
                          & T-junc-150-300 & 1.50 & 3.00 & 307 \\

    \bottomrule
    \end{tabular}}
    \label{tab:T-junction_dataset}
\end{table*}

\subsubsection{Parameter setting}
Similar to past studies on trajectory prediction \cite{9043898, 7780479, DBLP}, we use an 8-time-step lookback window to forecast the velocity at the next time step. \cite{doi:10.1086/200975} categorized the interpersonal distance into four zones: intimate (up to 0.46m), personal (0.46 to 1.2m), social (1.2 to 3.7m), and public (greater than 3.7m). In this study, the social interaction radius $R$ is considered to fall within the social distance category and is thus set as 1.2m. According to \cite{ZHAO2021103260}, the model's performance remains relatively stable and less sensitive when the central angle $\alpha$ is below 20$^\circ$. Therefore, we set $\alpha$ to 18$^\circ$, considering both model performance and computational efficiency. To evaluate the robustness of our model and explore appropriate parameter ranges, we vary two visual-information-related parameters: the virtual exit-distance parameter $D_e$ and the angle $\beta$. The parameter $D_e$ should be significantly greater than any distance between pedestrians and walls. Therefore, we evaluate $D_e$ at 20m, 30m, 50m and 100m, taking into account the dimensions of the experimental scenarios. The interval for the angle $\beta$ is evaluated at 5$^\circ$, 10$^\circ$, 15$^\circ$ and 18$^\circ$. This selection of values for $D_e$ and $\beta$ ensures us to cover a broad range, encompassing small to large distances and sparse to dense visual information. Details of the SVTCN hyperparameters can be found in Table \ref{tab:hyperparameters}. Specifically, consistent with \cite{bai2018empirical}, we use the Adam optimizer \cite{kingma2017adam} for training. The learning rate is set to 0.0001. 3000 iterations for each experiment were conducted to ensure convergence. The kernel size $q$ is set to 8 for an adequate receptive field. An exponential dilation $h = 2^{m-1}$ for the $m^{th}$ TCN residual block in the network is employed.

% All experiments are conducted on a desktop with Intel(R) Xeon(R) CPU E5-2620 v3 @ 2.40GHz, 2133MHz × 4 × 8GB RAM, and NVIDIA GeForce RTX 3080.

\begin{table}[htb]
    \centering
    \footnotesize
    \caption{SVTCN hyperparameters. In $(q, h, c)$, $q$, $h$ and $c$ denote the kernel size, dilation factor and channel of the 1D dilated convolutional operation,  respectively.}
    \begin{tabular}{
    >{\centering}m{0.3\textwidth} 
    >{\centering\arraybackslash}m{0.1\textwidth}}
      \toprule %Add a bold line to the header
      Hyperparameter & Value  \\
      \hline
      Optimizor & Adam \\
      Learning rate & 0.0001 \\
      Iteration & 3000 \\
      $(q, h, c)$ in the $1^{st}$ TCN residual block & $(8, 1, 32)$ \\
      $(q, h, c)$ in the $2^{nd}$ TCN residual block & $(8, 2, 64)$ \\
      $(q, h, c)$ in the $3^{rd}$ TCN residual block & $(8, 4, 96)$ \\ 
    \bottomrule
    \end{tabular}
    \label{tab:hyperparameters}
\end{table}

% \subsubsection{Rolling forecast setting}
% Upon the completion of model training, the trained model is used to conduct RF in the simulation scenarios. The simulation scenarios are designed based on the real-world testing scenarios (as indicated in Tables \ref{tab:corridor_dataset}-\ref{tab:T-junction_dataset}) to accurately evaluate the simulation performance in terms of realism. Specifically, the geometries of the simulation scenarios, number of participants, entry time of each pedestrian, and pedestrian positions during the initial $w$ (lookback window) time steps in the simulation scenarios are identical to those in the real-world testing experiments.

\subsection{Evaluation metrics}
Both qualitative and quantitative metrics are used to evaluate the performance of the proposed VID model in terms of realism. They are detailed below.
\subsubsection{Qualitative metrics}
\begin{enumerate}
\item Trajectories. Comparison of the crowd trajectories in the simulation and corresponding real-world scenarios is the most intuitive and effective approach for model evaluation. We define the trajectory point set for pedestrian $i$ in the real-world controlled experiment as $TRAJ_i^{expt} = \{\mathbf{p}_i^{1,expt}, \cdots, \mathbf{p}_i^{t,expt}, \cdots, \mathbf{p}_i^{T_i^{expt},expt}\}$. Here, $T_i^{expt}$ indicates the travel time of pedestrian $i$ in the controlled experiment, and $\mathbf{p}_i^{t,expt} \in \mathbb{R}^{1 \times 2}$ denotes the $xy$ coordinates of the trajectory point for pedestrian $i$ at time step $t$. Similarly, the trajectory point set for pedestrian $i$ in the simulation is defined as $TRAJ_i^{sim} = \{\mathbf{p}_i^{1,sim}, \cdots, \mathbf{p}_i^{t,sim}, \cdots, \mathbf{p}_i^{T_i^{sim},sim}\}$, where $T_i^{sim}$ is the travel time of pedestrian $i$ in the simulation. The number of pedestrians is $N_p$.

\item Fundamental diagram. The fundamental diagram, which is a widely used tool in the field of pedestrian and evacuation dynamics, describes the relationship between the velocity/flow and density. In this study, we use Voronoi density and velocity (refer to the approach described in \cite{STEFFEN20101902}). Specifically, the flow $J$ of the measurement area is defined as

\begin{equation}
J = \rho vb
\end{equation}

where $v$, $\rho$ and $b$ denote the Voronoi velocity, density and width of the measurement area, respectively .

\item Density-time and velocity-time profiles. Density and velocity are fundamental attributes characterizing pedestrian motion. Consistency in velocity and density changes over time in controlled experiments and simulations provide evidence for the effectiveness of the model. The Voronoi density and velocity are also used in this evaluation.
\end{enumerate}

\subsubsection{Quantitative metrics}
\begin{enumerate}

\item Egress time error (ETE) and percentage egress time error (PETE). Egress time represents the time interval between the first pedestrian entering the scenario and the last pedestrian exiting the scenario. This metric is widely used in safety evacuation research, especially for determining the RSET. To calculate ETE and PETE, we denote the time steps of the first pedestrian entering the scenario in the controlled experiment and simulation as $T^{expt, start}$ and $T^{sim, start}$, respectively. Similarly, the time steps of the last pedestrian exiting the scenario in the controlled experiment and simulation are denoted as $T^{expt, end}$ and $T^{sim, end}$, respectively. The ETE is defined as the absolute difference between the egress times in the simulation and controlled experiment:

\begin{equation}
ETE = |(T^{sim, end} - T^{sim, start}) - (T^{expt, end} - T^{expt, start})|
\end{equation}
The PETE is defined as the ratio of the ETE to the egress time in the controlled experiment:
\begin{equation}
PETE = \frac{ETE}{T^{expt, end} - T^{expt, start}}
\end{equation}

\item Travel time error (TTE) and percentage travel time error (PTTE). TTE indicates the difference between the simulated and actual travel time. PTTE represents this difference as a percentage of the actual travel time. The TTE and PTTE for each pedestrian $i$ are defined as follows:
\begin{equation}
TTE_i = |T_i^{sim} - T_i^{expt}|
\end{equation}

\begin{equation}
PTTE_i = \frac{TTE_i}{T_i^{expt}}
\end{equation}

\item Trajectory displacement error (TDE) and final displacement error (FDE). 
TDE quantifies the mean displacement error of a pedestrian throughout their travel between controlled experiments and simulations. Given that the simulated travel time $T_i^{sim}$ and actual travel time $T_i^{expt}$ may differ, the approach proposed by \cite{ZHAO2021103260} is used to define $TDE_i$:
\begin{equation}
TDE_i = \frac{\sum_{t=1}^{T_i^{expt}} 
\min \limits_{\tau \in \{1,\cdots,T_i^{sim}\}} \| \mathbf{p}_i^{\tau,sim} - \mathbf{p}_i^{t,expt}\|}{T_i^{expt}} 
\end{equation}

FDE measures the displacement error of a pedestrian at the end of their travel, specifically when they exit the scenario. $FDE_i$ is defined as the Euclidean distance between the actual position $\mathbf{p}_i^{T_i^{expt},expt}$ and simulated position $\mathbf{p}_i^{T_i^{sim},sim}$ of pedestrian $i$ at the end of their travel:
\begin{equation}
FDE_i = ||\mathbf{p}_i^{T_i^{expt},expt} - \mathbf{p}_i^{T_i^{sim},sim}||
\end{equation}

\end{enumerate}

\section{Results}

\subsection{Parameter Sensitivity}
\begin{figure*}[htpb]
\centering
\includegraphics[width=0.98\textwidth]{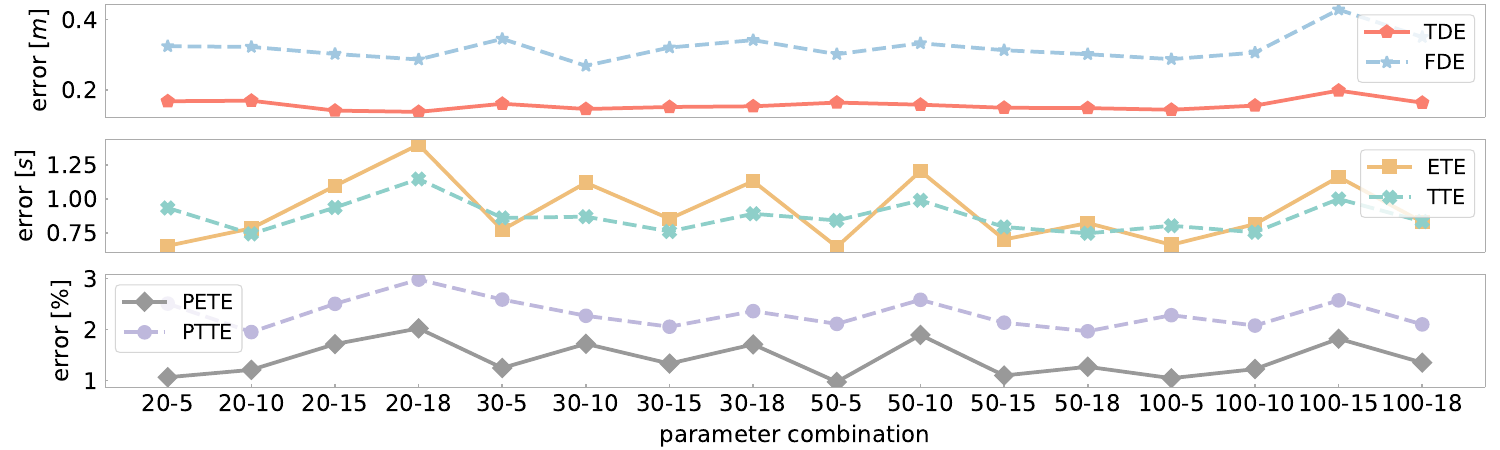}
\caption{Values of quantitative metrics (TDE, FDE, ETE, TTE, PETE and PTTE) for the 16 parameter combinations tested. The x-axis labels represent the $D_e$-$\beta$ combinations. For instance, 20-5 indicates the case where $D_e=20m$ and $\beta=5^\circ$. Each data point is calculated using results from all three geometries (corridor, corner and T-junction).}
\label{fig:param_senti}
\end{figure*}

As described in Section \ref{sec:experiments}, parameter sensitivity tests regarding the virtual exit-distance parameter $D_e$ and the angle parameter $\beta$ were conducted to assess model robustness and explore suitable parameter ranges. Specifically, we evaluated $D_e$ at distances of 20m, 30m, 50m and 100m, as well as $\beta$ at angles of 5$^\circ$, 10$^\circ$, 15$^\circ$ and 18$^\circ$. Consequently, a total of 16 (4 $\times$ 4, 4 choices for $D_e$ and 4 choices for $\beta$) parameter combinations were tested. Fig. \ref{fig:param_senti} presents the results of the six quantitative metrics (FDE, TDE, TTE, ETE, PTTE and PETE) for these 16 combinations. Each data point in Fig. \ref{fig:param_senti} represents the average value obtained from all samples within the 16 testing scenarios (6 for corridor, 6 for corner, and 4 testing scenarios for T-junction geometries). Observations from Fig. \ref{fig:param_senti} indicate that our model exhibits robustness and insensitivity within the parameter ranges of $D_e$ (20m to 100m) and $\beta$ ($5^\circ$ to $18^\circ$). This is evidenced by the consistent stability observed across all six quantitative metrics for the 16 $D_e$-$\beta$ combinations. Specifically, almost all the differences in TDE/FDE values among the 16 parameter combinations are below 0.1m, while the differences in ETE/TTE and PETE/PTTE values are below 0.5s and 1\%, respectively.

\subsection{Qualitative and quantitative results}
In order to demonstrate the superiority of our VID model, we conducted qualitative comparisons with controlled experiments and the previous state-of-the-art model, DCLN \cite{9043898}. Additionally, we performed quantitative error comparisons with DCLN. DCLN is a deep convolutional LSTM network. It employs discrete grids with a size of 5$\times$5 units, where each grid measures 0.4m in both length and width, to extract social and spatial information. Specifically, a total of six 5$\times$5 grids are used to capture the velocities of subject pedestrians and their neighbours, the relative positions of their neighbours, the existence of obstacles and walls around subject pedestrians, and the relative position of the exit. Empirical experiments conducted by Song et al. \cite{9043898} have demonstrated the remarkable performance of DCLN, surpassing both the canonical SF model \cite{Helbing2000} and Social-LSTM \cite{7780479} in room and corridor scenarios. Furthermore, their study shows that DCLN outperforms a range of classical pedestrian trajectory prediction data-driven models, including Social-LSTM \cite{7780479}, SS-LSTM \cite{08bba04d17ad491abd48adb734a4fadd}, Social-GAN \cite{Social-GAN} and Scene-LSTM \cite{Scene-LSTM}, in terms of average results on five widely-used public pedestrian datasets. These findings affirm the effectiveness of DCLN in capturing pedestrian dynamics and highlight its strong capabilities. Consequently, in this study, we compare our model with DCLN. We reproduce DCLN with the same training-validation and testing dataset (Tables \ref{tab:corridor_dataset}-\ref{tab:T-junction_dataset}). Notably, since our model are insensitive within the parameter ranges, we use the results from the previously mentioned 16 experiments with different $D_e$-$\beta$ combinations to calculate the quantitative metrics (i.e., ETE, PETE, TTE, PTTE, TDE and FDE) for the VID model to ensure a fairer comparison.

\begin{figure*}[htpb]
    \centering
    \includegraphics[width=0.98\textwidth]{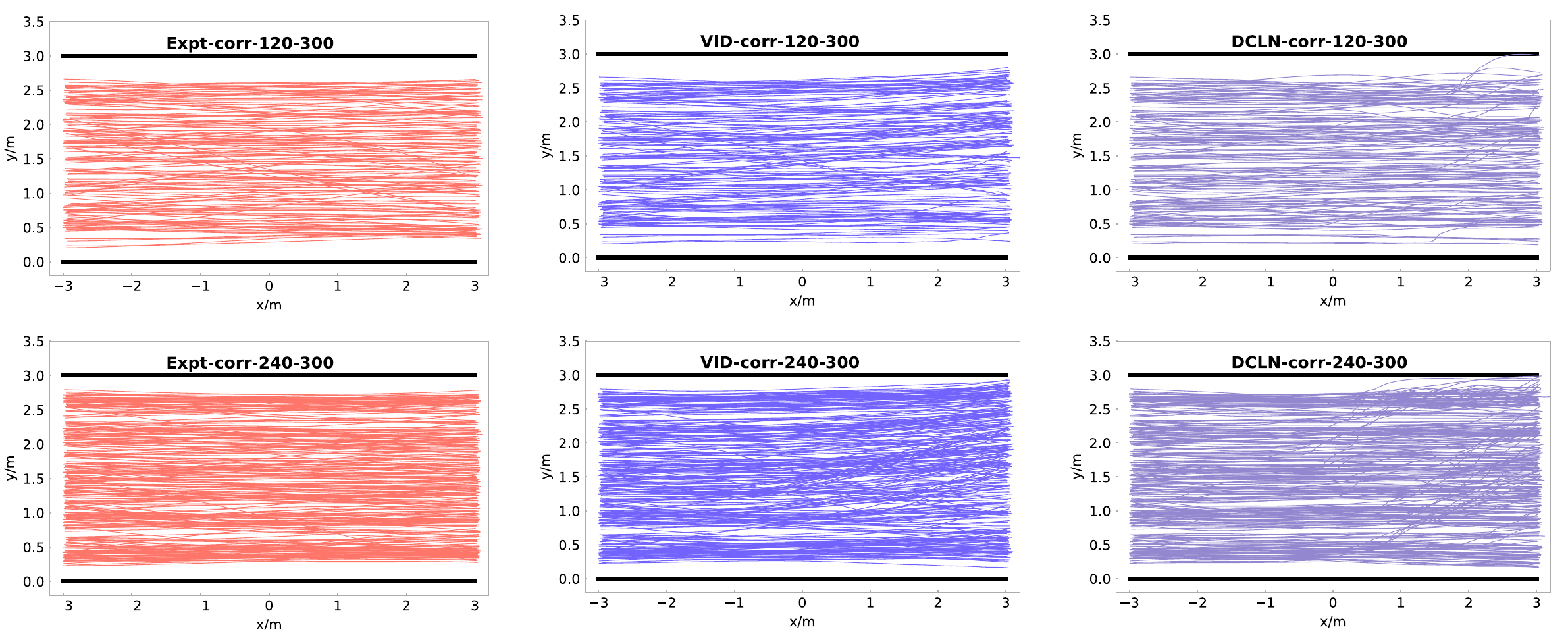}
    \caption{Trajectories obtained from experiments, our VID model and DCLN \cite{9043898} in the corridor scenario of corr-120-300 and corr-240-300.}
    \label{fig:trajs_uni}
    \end{figure*}

\begin{figure*}[htpb]
    \centering
    \includegraphics[width=0.98\textwidth]{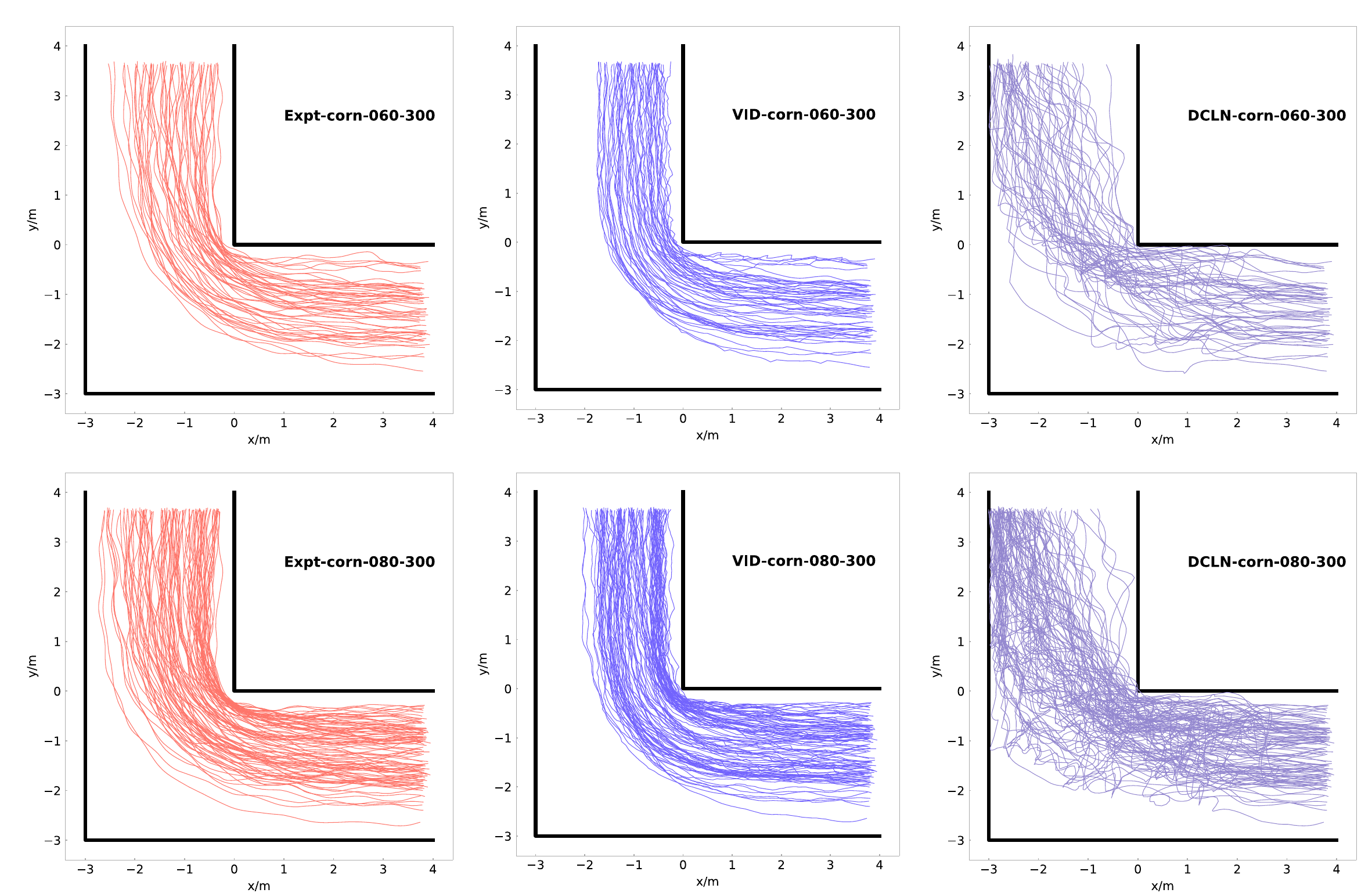}
    \caption{Trajectories obtained from experiments, our model and DCLN \cite{9043898} in the corner scenario of corn-060-300 and corn-080-300.}
    \label{fig:trajs_corner}
    \end{figure*}

\begin{figure*}[htpb]
    \centering
    \includegraphics[width=0.98\textwidth]{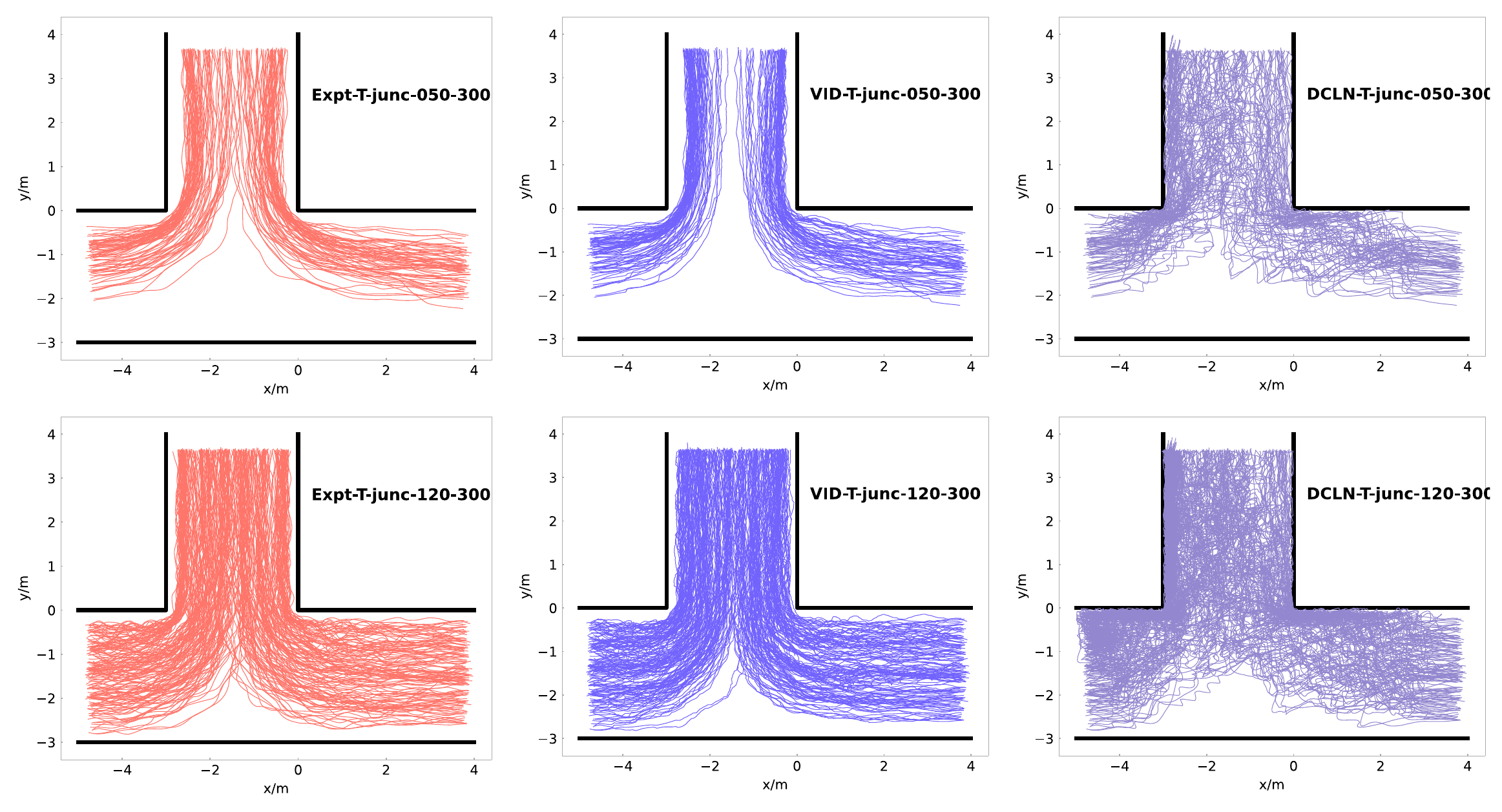}
    \caption{Trajectories obtained from experiments, our model and DCLN \cite{9043898} in the T-junction scenario of T-junc-050-300 and T-junc-120-300.}
    \label{fig:trajs_T-junc}
    \end{figure*}

\textbf{Trajectories}. Figs. \ref{fig:trajs_uni}-\ref{fig:trajs_T-junc} present the trajectories obtained from the controlled experiments, our model with parameters $D_e=100m$ and $\beta=5^\circ$, as well as DCLN \cite{9043898}, for the corridor, corner, and T-junction geometries, respectively. The simulated trajectories from our model exhibit a remarkable resemblance to the actual trajectories observed in the controlled experiments. Moreover, our model demonstrates superior performance in terms of trajectory realism compared to DCLN.

% \begin{figure}[htpb]
    % \centering
    % % \setlength{\abovecaptionskip}{-0.0005cm}
    % \includegraphics[width=0.98\textwidth]{images/analysis/trajs/uni_corridor.pdf}
    % \caption{Experimental and simulated trajectories for different runs in corridor geometry.}
    % \label{fig:trajs_uni}
    % \end{figure}

    % \begin{figure}[htpb]
    % \centering
    % % \setlength{\abovecaptionskip}{-0.0005cm}
    % \includegraphics[width=0.98\textwidth]{images/analysis/trajs/corner.pdf}
    % \caption{Experimental and simulated trajectories for different runs in corner geometry.}
    % \label{fig:trajs_corner}
    % \end{figure}
    
    % \begin{figure}[htpb]
    % \centering
    % % \setlength{\abovecaptionskip}{-0.0005cm}
    % \includegraphics[width=0.98\textwidth]{images/analysis/trajs/T-junction.pdf}
    % \caption{Experimental and simulated trajectories for different runs in T-junction geometry.}
    % \label{fig:trajs_T-junc}
    % \end{figure}

\begin{figure*}[htpb]
    \centering
    \includegraphics[width=0.98\textwidth]{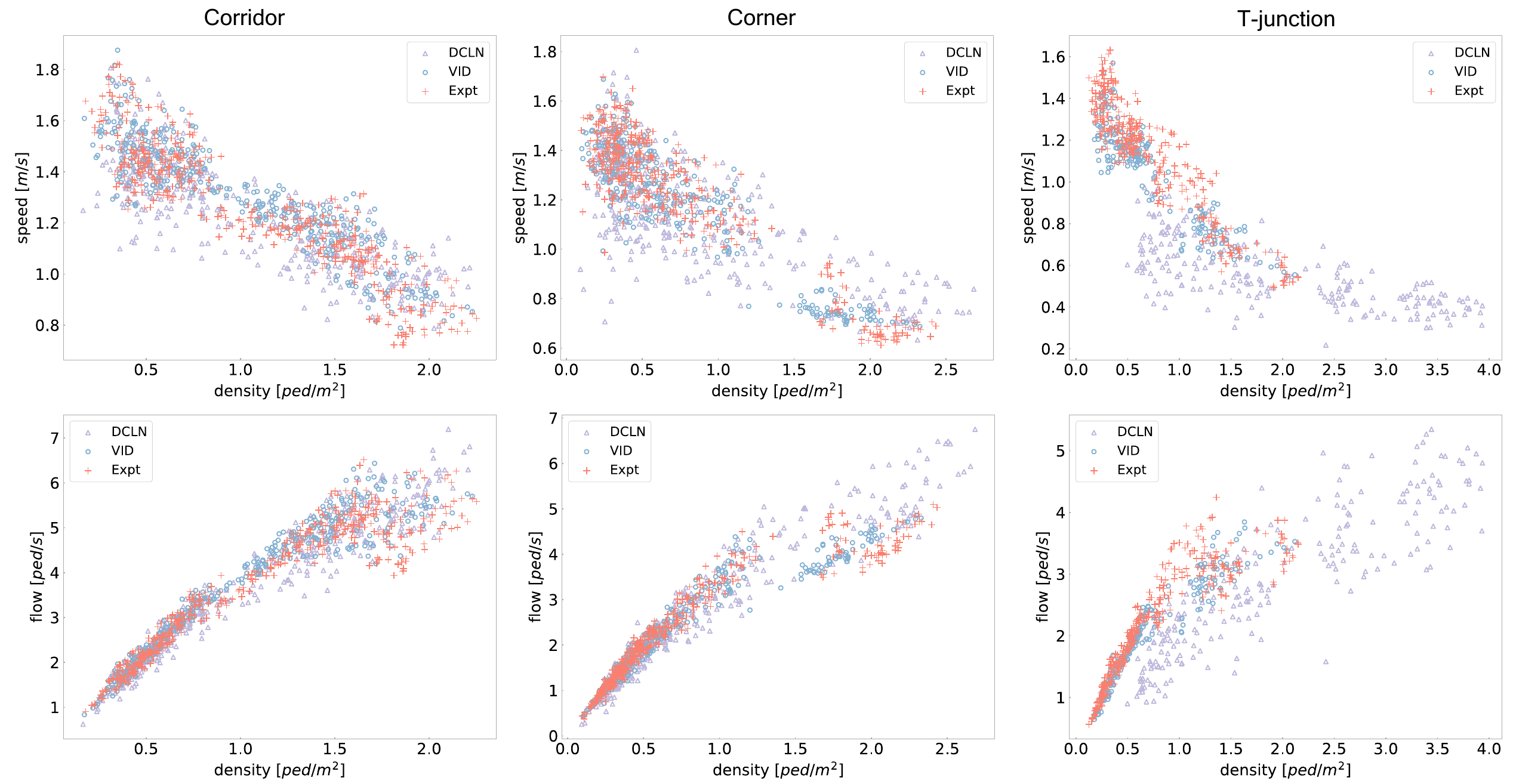}
    \caption{Fundamental diagrams derived from controlled experiments, our VID model and DCLN \cite{9043898} for the corridor, corner and T-junction geometries, respectively.}
    \label{fig:fundamental_diagram}
    \end{figure*}

\textbf{Fundamental diagram}. The measurement areas used for computing the fundamental diagrams for the three geometries are depicted in Fig. \ref{fig:exp_setup}. The trajectory data from all testing scenarios, as presented in Tables \ref{tab:corridor_dataset}-\ref{tab:T-junction_dataset}, are utilized to generate the fundamental diagram for each respective geometry. Fig. \ref{fig:fundamental_diagram} presents the fundamental diagrams obtained from the controlled experiments, our VID model with parameters $D_e=100m$ and $\beta=5^\circ$, as well as DCLN \cite{9043898}. The fundamental diagrams derived from our VID model exhibit a high degree of agreement with those generated from the experimental data. Furthermore, our model outperforms DCLN in term of fundamental diagrams due to its significantly smaller deviation from the experimental data compared to DCLN.

\begin{figure*}[htpb]
    \centering
    \includegraphics[width=0.98\textwidth]{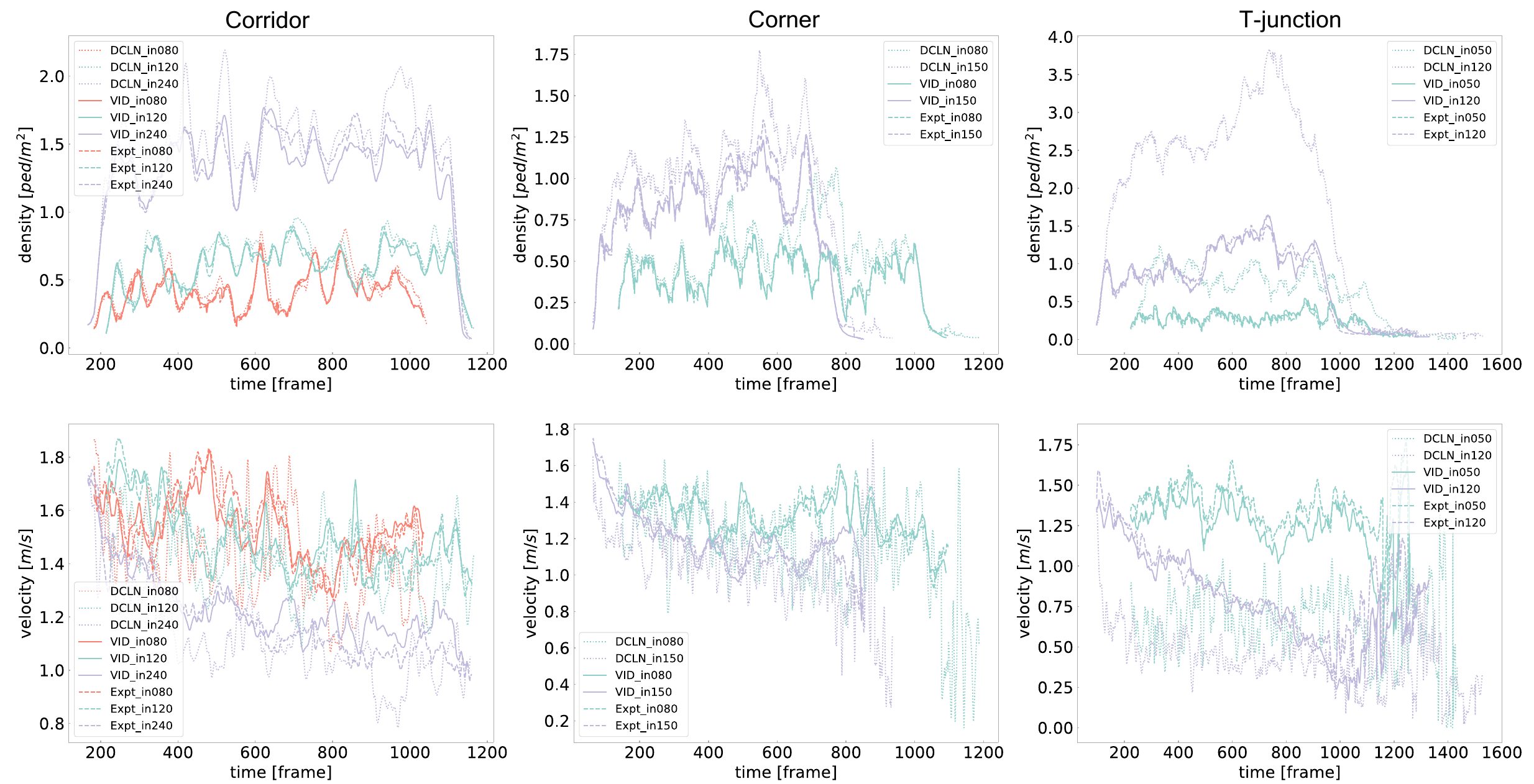}
    \caption{Density-time and velocity-time profiles obtained in controlled experiments, our VID model and DCLN \cite{9043898} for corridor, corner and T-junction geometries.}
    \label{fig:time_v_rho}
    \end{figure*}

\textbf{Density-time and velocity-time profiles}. Fig. \ref{fig:time_v_rho} shows the density and velocity variations over time in the controlled experiments, our VID model with parameters $D_e=100m$ and $\beta=5^\circ$, as well as DCLN \cite{9043898} for the corridor, corner and T-junction geometries. The density and velocity correspond to the Voronoi density and velocity of the measurement area, respectively. A subset of the test scenarios is presented in the figure to enhance clarity. The density-time and velocity-time plots obtained from the controlled experiments and our VID  model are highly consistent in terms of both the magnitude and trend. Moreover, it is evident that our model surpasses DCLN in capturing more realistic density and velocity variations.

\textbf{ETE and PETE}. Table \ref{tab:ETE_PETE} presents the ETE and PETE results obtained from our VID model and DCLN \cite{9043898} for each test scenario in the corridor, corner and T-junction geometries. The values reported in the table for the VID model represent the median values obtained from the 16 experiments with different $D_e$-$\beta$ combinations. One can see that our VID model exhibits high accuracy in terms of ETE and PETE, as evidenced by the PETE values ranging from $0.10\%$ to $4.27\%$ across all test scenarios. These findings suggest that our model could serve as a potential alternative to RSET calculation. Moreover, our model demonstrates superior performance overall compared to DCLN regarding ETE and PETE across all three geometries.

    % \begin{table*}[htb]
    % \centering
    % \footnotesize
    % \caption{ETE and PETE values for each scenario of the corridor, corner, and T-junction geometries.}
    %     \resizebox{1.02\textwidth}{!}{
    % \begin{tabular}{lcccccccccccccccccc}
    %   \toprule %Add a bold line to the 
    
    %   & \multicolumn{6}{c}{Corridor} & & \multicolumn{6}{c}{Corner} & & \multicolumn{4}{c}{T-junction}\\
    %                      \cline{2-7} \cline{9-14} \cline{16-19}
    %          $b_{in}[cm]$          & 80 & 100 & 120 & 180 & 240 & 300 & & 50 & 60 & 80 & 100 & 150 & 300 & & 50 & 80 & 120 &150\\
    % \hline
    % ETE$[s]$ & 0.19 & 0.25 & 0.31 & 0.25 & 0.75 & 2.00 & & 0.25 & 0.38 & 0.50 & 0.25 & 2.13 & 2.94 & & 0.06 & 0.50 & 2.06 &1.94\\
    % PETE$[\%]$ & 0.33 & 0.52 & 0.50 & 0.45 & 1.18 & 2.63 & & 0.42 & 0.80 & 0.79 & 0.42 & 4.12 & 4.62 & & 0.09 & 0.72 & 2.62 &2.52\\
    % \bottomrule
    % \end{tabular}    
    % }
    % \label{tab:ETE_PETE}
    % \end{table*}

    \begin{table*}[htb]
    \centering
    \footnotesize
    \caption{ETE and PETE values obtained from our VID model and DCLN \cite{9043898} for each scenario of the corridor, corner and T-junction geometries.}
        \resizebox{1.02\textwidth}{!}{
    \begin{tabular}{lccccccccccccccccccc}
      \toprule %Add a bold line to the 
    
      & & \multicolumn{6}{c}{Corridor} & & \multicolumn{6}{c}{Corner} & & \multicolumn{4}{c}{T-junction}\\
                         \cline{3-8} \cline{10-15} \cline{17-20}
             & $b_{in}[cm]$          & 80 & 100 & 120 & 180 & 240 & 300 & & 50 & 60 & 80 & 100 & 150 & 300 & & 50 & 80 & 120 &150\\
    \hline
    \multirow{2}*{ETE$[s]$}
    & VID & 0.09 & 0.13 & 0.06 & 0.50 & 0.47 & 1.09 & & 0.38 & 0.53 & 0.63 & 1.22 & 1.50 & 2.72 & & 0.18 & 0.25 & 1.84 &2.40\\
    & DCLN & 1.25 & 0.63 & 0.06 & 0.25 & 1.12 & 2.18 & & 3.00 & 4.00 & 6.56 & 4.87 & 8.44 & 6.88 & & 9.38 & 14.69 & 14.13 &13.75\\
    % \hline
    \hline
    \multirow{2}*{PETE$[\%]$}
    & VID & 0.16 & 0.26 & 0.10 & 0.90 & 0.73 & 1.44 & & 0.62 & 1.13 & 0.98 & 2.03 & 2.90 & 4.27 & & 0.27 & 0.36 & 2.34 &3.12\\
    & DCLN & 2.18 & 1.30 & 0.10 & 0.45 & 1.76 & 2.88 & & 5.02 & 8.51 & 10.36 & 8.13 & 16.34 & 10.82	& & 13.55 & 21.23 & 17.97 & 17.87 \\
    \bottomrule
    \end{tabular}    
    }
    \label{tab:ETE_PETE}
    \end{table*}

\begin{figure*}[htpb]
    \centering
    \includegraphics[width=0.98\textwidth]{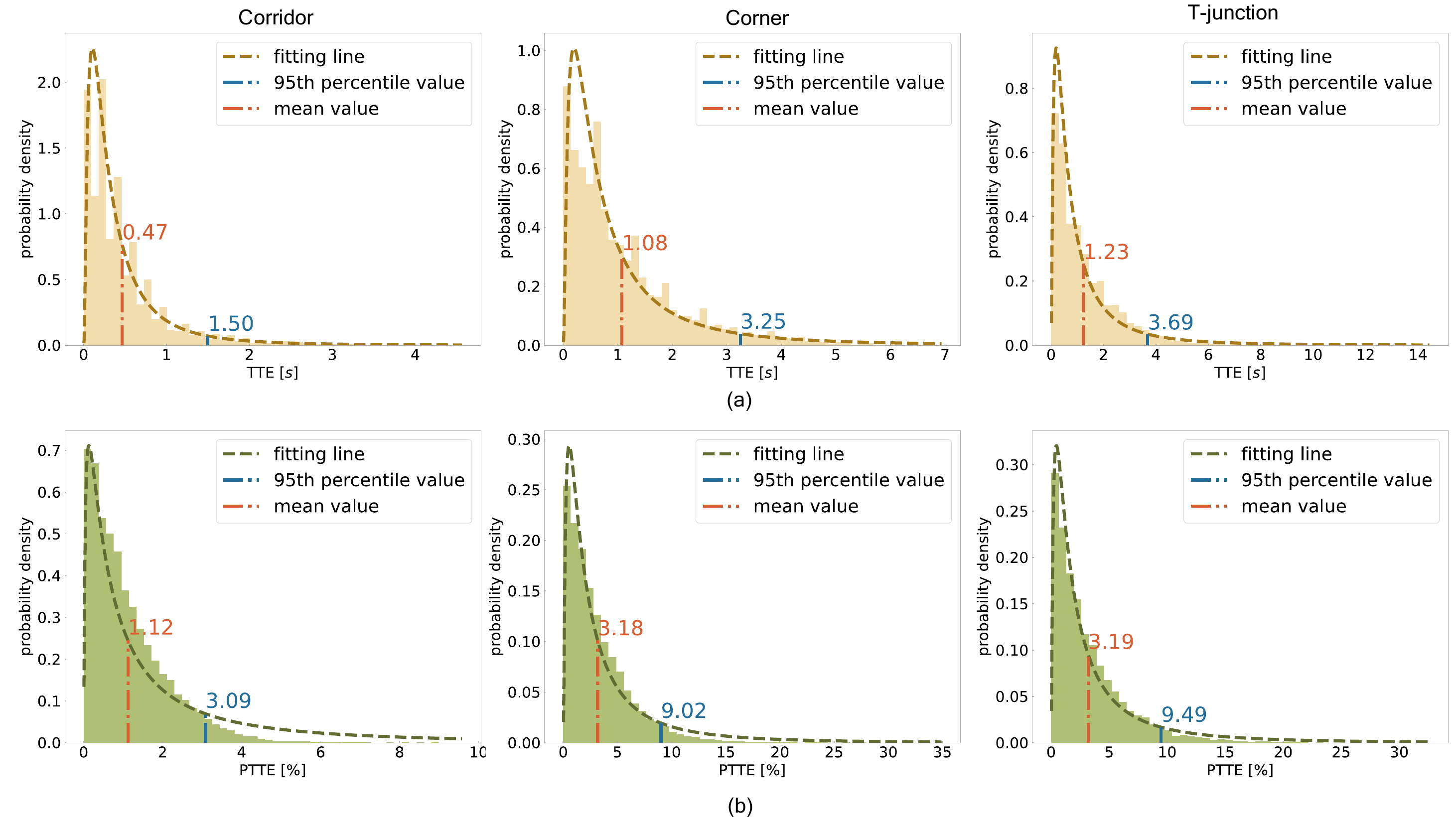}
    \caption{Probability density distributions of (a) TTE and (b) PTTE obtained from our VID model for the corridor, corner and T-junction geometries. Both TTE and PTTE for each geometry follow log-normal distributions. The orange and blue dash-dotted lines indicate the mean and 95th percentile values, respectively.}
    \label{fig:TTE_PTTE}
    \end{figure*}

\textbf{TTE and PTTE}. The probability density distributions of TTE and PTTE obtained from the total 16 experiments with different $D_e$-$\beta$ combinations in our VID model are shown in Figs. \ref{fig:TTE_PTTE}(a) and \ref{fig:TTE_PTTE}(b), respectively. The proposed model achieves a satisfactory performance in reflecting the pedestrian travel time, as indicated by the small values of TTE and PTTE. Specifically, the mean TTE for all three geometries is less than $1.23s$, with 95th percentile values smaller than $3.69s$. Moreover, the mean PTTE is less than $3.19\%$, with 95th percentile values smaller than $9.49\%$ for all geometries. Table \ref{tab:TTE_PTTE_TDE_FDE} presents the mean values of TTE and PTTE obtained from our VID model and DCLN \cite{9043898} for the corridor, corner and T-junction geometries. The results demonstrate that the travel times predicted by our model exhibit greater consistency with those observed in the controlled experiments across all three geometries, in comparison to DCLN.

\begin{table}[htb]
    \centering
    \scriptsize
    \caption{Mean values of TTE, PTTE, TDE and FDE obtained from our VID model and DCLN \cite{9043898} for corridor, corner and T-junction geometries.}
    \begin{tabular}{    
    >{\centering}m{0.07\textwidth} 
    >{\centering\arraybackslash}m{0.07\textwidth}
    >{\centering\arraybackslash}m{0.07\textwidth}
    >{\centering\arraybackslash}m{0.07\textwidth}
    >{\centering\arraybackslash}m{0.07\textwidth}}
      \toprule %Add a bold line to the 
      & & Corridor &Corner & T-junction \\
    \hline
    \multirow{2}*{TTE$[s]$}
    & VID & 0.47 & 1.08 & 1.23 \\
    & DCLN & 0.64 & 5.27 & 12.95 \\
    % \hline
    \hline
    \multirow{2}*{PTTE$[\%]$}
    & VID & 1.12 & 3.18 & 3.19 \\
    & DCLN & 1.72 & 17.84 & 38.56 \\
        \hline
    \multirow{2}*{TDE$[m]$}
    & VID & 0.10 & 0.24 & 0.17 \\
    & DCLN & 0.09 & 0.49 & 0.40 \\
    % \hline
    \hline
    \multirow{2}*{FDE$[m]$}
    & VID & 0.25 & 0.49 & 0.29 \\
    & DCLN & 0.17 & 1.15 & 0.74 \\
    \bottomrule
    \end{tabular}    
    \label{tab:TTE_PTTE_TDE_FDE}
    \end{table}    

\begin{figure*}[htpb]
    \centering
    \includegraphics[width=0.98\textwidth]{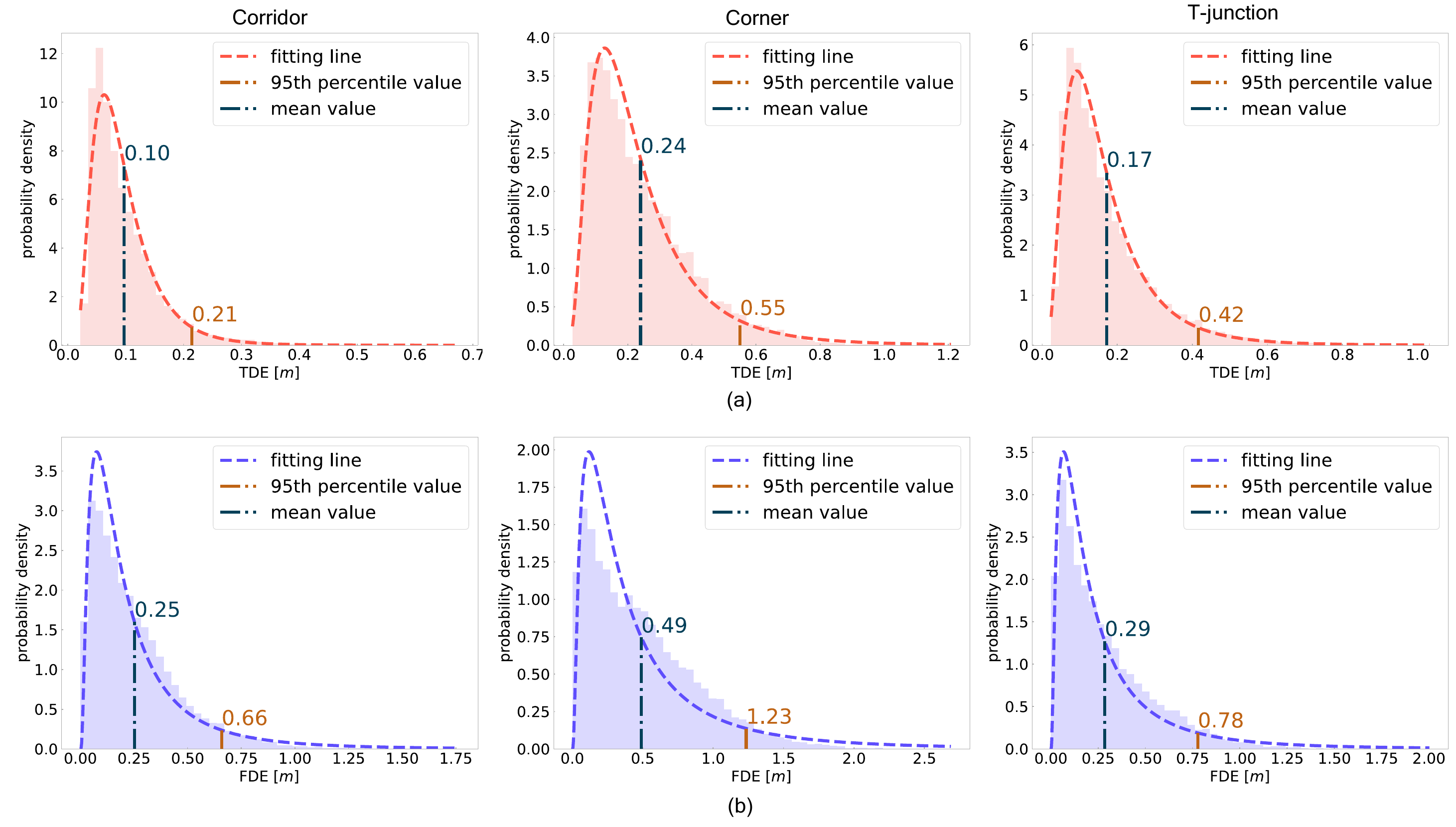}
    \caption{Probability density distributions of (a) TDE and (b) FDE obtained from our VID model for the corridor, corner and T-junction geometries. Both TDE and FDE for each geometry follow log-normal distributions. The blue and yellow dash-dotted lines indicate the mean and 95th percentile values, respectively.}
    \label{fig:TDE_FDE}
    \end{figure*}

\textbf{TDE and FDE}. Fig. \ref{fig:TDE_FDE} shows the TDE and FDE distributions obtained from the 16 experiments with different $D_e$-$\beta$ combinations in our VID model for each geometry. The mean TDE values for the corridor, corner and T-junction geometries are $0.10m$, $0.24m$ and $0.17m$, respectively, whereas the 95th percentile values are $0.21m$, $0.55m$ and $0.42m$, respectively. These results indicate that the trajectory error is generally small, particularly in the corridor and T-junction geometries. Similarly, the FDE exhibits small values in the corridor and T-junction geometries, with the mean values being $0.25m$ and $0.29m$, respectively, which do not exceed $10\%$ of the corridor width ($3m$). Table \ref{tab:TTE_PTTE_TDE_FDE} provides a comparison of the mean TDE and FDE values obtained from our VID model and DCLN \cite{9043898} for the corridor, corner and T-junction geometries. While DCLN outperforms the VID model slightly in terms of TDE and FDE for the corridor geometry, the latter demonstrates clear superiority over the former in the case of corner and T-junction geometries.

\section{discussion}
The qualitative and quantitative results demonstrate that the proposed model performs well across various geometries. This observation reflects the improved adaptability of the proposed VID model and effectiveness of the RGL method. Previous research on data-driven crowd simulations has predominantly focused on considering only one specific geometry \cite{ZHAO2020123825, 7464842, ZHAO2021103260, 9898931}, resulting in models that lack the ability to be applied to various geometries. We posit that this limitation is due to the narrow focus on influencing factors specific to the geometry being studied. For example, in the work of \cite{7464842}, the geometry information was restricted to only three boundary attributes: distance to the left boundary, distance to the right boundary, and distance to the target line. While such simplified boundary representations may suffice for their study of a crosswalk, they may prove inadequate for more complex geometries such as corners, T-junctions or bottlenecks. 
\par
% In this regard, traditional social force model can serve as a point of reference, as they capture the most universal rules and can be employed in almost all geometries.
Despite the diversity of environments and geometries, we believe that certain universal factors govern pedestrian motion. By comprehensively considering and effectively extracting these factors, we can build data-driven models that are adaptive to a wider range of geometries. In this context, a key factor is visual information, including geometry and locomotion data. When pedestrians navigate, they plan their intended global route based on the scenario geometry and their locomotion within their vision field. Here, the global route refers to the route from the current position of a pedestrian to their target position within the vision field. Concurrently, local pedestrian motion is influenced by their previous motion and interactions with neighbouring pedestrians, such as following or avoidance. In essence, we posit that geometry-locomotion information largely determines a pedestrian’s global route, whereas social information and previous motion influence their local behaviour at every time step.

\par

In the SF model, a pedestrian’s desired velocity propels their motion, and the direction of this velocity primarily determines their intended global route. Conventionally, a single point serves as the target for each pedestrian to determine their desired direction. However, this simplistic method may not be adequate for certain cases, such as corner geometries. Instead of using a single point, we introduce more complex geometry-locomotion information to navigate pedestrians.
\par
The stable performance of the quantitative metrics in the parameter sensitivity tests regarding $D_e$ and $\beta$ demonstrates the robustness of our VID model. Moreover, the results indicating insensitivity to variations in $D_e$ at distances of 20m, 30m, 50m and 100m, as well as $\beta$ at angles of 5$^\circ$, 10$^\circ$, 15$^\circ$ and 18$^\circ$, highlight the existence of a wide safe zone for selecting parameter values. These findings can also provide reference values for the selection of model parameters in future research and practical applications.
\par

Overall, the proposed VID model demonstrates superior performance over DCLN \cite{9043898} in both qualitative and quantitative metrics across all three scenarios. This superiority may be attributed to two factors, namely the global and local aspects. From a global perspective, DCLN solely relies on the input of the terminal point position, failing to provide information regarding the scenario geometry and exit width, which impedes the astute planing of global routes. From a local standpoint, DCLN employs discrete grids that can only indicate the presence of pedestrians or walls within a given grid area, without providing precise positional details. This imprecise representation hinders the effective determination of local pedestrian collision avoidance interactions with surrounding entities.

\par 
% We evaluate our approach on three public datasets including corridor, corner, and T-junction geometries. Our results, both qualitative and quantitative, indicate that our model performs well. However, our work has certain limitations. Firstly, while we utilized the same parameters for the models of the three aforementioned geometries, this does not necessarily mean that these parameters would be suitable for other geometries such as bottleneck. For example, pedestrians may experience significant blockage in bottleneck scenarios, leading to frequent backward movement. In such situations, their visual field may not be aligned with their velocity direction. Therefore, it may be necessary to use another approach to define the forward motion zone. Secondly, we did not account for obstacles in our model. In most prior studies, obstacles are considered to be local influence factors. That is to say, they are only taken into account when pedestrians are in close proximity to them. However, obstacles may also significantly impact a pedestrian's intended global route, even if they are far away. Hence, obstacles in data-driven crowd simulation deserve more detailed research. 

The main contribution of our work is the visual information-driven navigation and the RGL method. Experimental results indicate the enhanced adaptability of the proposed VID model and the effectiveness of the RGL method. Our research holds both theoretical and practical significance. First, the incorporation of visual information enables us to achieve superior adaptability, thereby serving as a valuable source of inspiration for future studies in this domain. Second, the more realistic crowd simulation generated by our model finds applications in various fields such as facility design, gaming and animation. Third, the results of the comprehensive parameter sensitivity tests provide practical guidance for parameter selection. Last, our model can serve as a potential alternative for calculating RSET in the field of fire safety due to its ability to generate low ETE values.

\par However, it is important to acknowledge the limitations of our study. First, the current proposed model may not be well-suited for situations characterized by frequent backward movement, such as bottlenecks with significantly high pedestrian densities. In such cases, the presence of substantial congestion often leads to frequent instances of pedestrian backward movement, resulting in frequent misalignment between pedestrians' visual field and their velocity direction. Thus, it may be necessary to explore alternative approaches to define the forward motion zone in such scenarios. Second, the validation of the model in complex scenarios has not been extensively explored within the scope of this paper. Future research efforts can delve into investigating the utilization and applicability of the model in various complex structures.

\section{Conclusion} 
This paper introduces a novel VID model to enhance the adaptability and realism of data-driven crowd simulations. A TCN-based neural network, named SVTCN, is incorporated into the VID model to predict pedestrian velocities based on their previous social-visual information and motion data. Additionally, the RGL method is established to extract the visual information, including scenario geometry and pedestrian locomotion. Three public pedestrian motion datasets with distinct geometries (i.e., corridor, corner and T-junction) are used to validate the proposed model. Rolling prediction is conducted in simulation scenarios, and the obtained simulation results are compared with the corresponding experimental data as well as those generated by a previous advanced model, DCLN \cite{9043898}. Qualitative and quantitative metrics are employed to evaluate the model performance. The results demonstrate the excellent performance of the proposed model across the three geometries, highlighting the effectiveness of the VID navigation method and adaptability of the proposed model. Moreover, the proposed VID model outperforms DCLN \cite{9043898} in both qualitative and quantitative metrics across all three scenarios. These findings provide valuable insights into the effectiveness of incorporating visual information into data-driven approaches. The proposed model exhibits promising potential for diverse applications, including facility design and evacuation guidance.

\par
Future research endeavours can focus on two key aspects. First, novel methods can be used to define the forward motion zone in scenarios involving significant backward movement. Second, the validation of the model in various complex structures can be investigated. These efforts would further advance the development of data-driven models and facilitate their practical implementation in real-world applications.

% \section*{APPENDIX}he validation of the model in complex structures can be investigated.
% Appendixes, if needed, appear before the acknowledgment.

\section*{ACKNOWLEDGMENT}
The work described in this paper was fully supported by the grants from CityU Project No. 7005769 and 7005895.

\section*{REFERENCES AND FOOTNOTES}
\bibliographystyle{IEEEtran}
\bibliography{ref}

\begin{thebibliography}{10}
\providecommand{\url}[1]{#1}
\csname url@samestyle\endcsname
\providecommand{\newblock}{\relax}
\providecommand{\bibinfo}[2]{#2}
\providecommand{\BIBentrySTDinterwordspacing}{\spaceskip=0pt\relax}
\providecommand{\BIBentryALTinterwordstretchfactor}{4}
\providecommand{\BIBentryALTinterwordspacing}{\spaceskip=\fontdimen2\font plus
\BIBentryALTinterwordstretchfactor\fontdimen3\font minus
  \fontdimen4\font\relax}
\providecommand{\BIBforeignlanguage}[2]{{%
\expandafter\ifx\csname l@#1\endcsname\relax
\typeout{** WARNING: IEEEtran.bst: No hyphenation pattern has been}%
\typeout{** loaded for the language `#1'. Using the pattern for}%
\typeout{** the default language instead.}%
\else
\language=\csname l@#1\endcsname
\fi
#2}}
\providecommand{\BIBdecl}{\relax}
\BIBdecl

\bibitem{Helbing2000}
D.~Helbing, I.~Farkas, and T.~Vicsek, ``Simulating dynamical features of escape
  panic,'' \emph{Nature}, vol. 407, no. 6803, pp. 487--490, Sep 2000.

\bibitem{VARAS2007631}
A.~Varas, M.~Cornejo, D.~Mainemer, B.~Toledo, J.~Rogan, V.~Muñoz, and
  J.~Valdivia, ``Cellular automaton model for evacuation process with
  obstacles,'' \emph{Physica A: Statistical Mechanics and its Applications},
  vol. 382, no.~2, pp. 631--642, 2007.

\bibitem{10.1007/3-540-28091-X_36}
S.~Hoogendoorn and W.~Daamen, ``Self-organization in pedestrian flow,'' in
  \emph{Traffic and Granular Flow '03}, S.~P. Hoogendoorn, S.~Luding, P.~H.~L.
  Bovy, M.~Schreckenberg, and D.~E. Wolf, Eds.\hskip 1em plus 0.5em minus
  0.4em\relax Berlin, Heidelberg: Springer Berlin Heidelberg, 2005, pp.
  373--382.

\bibitem{Helbing_1995}
D.~Helbing and P.~Moln{\'{a}}r, ``Social force model for pedestrian dynamics,''
  \emph{Physical Review E}, vol.~51, no.~5, pp. 4282--4286, may 1995.

\bibitem{9043898}
X.~Song, K.~Chen, X.~Li, J.~Sun, B.~Hou, Y.~Cui, B.~Zhang, G.~Xiong, and
  Z.~Wang, ``Pedestrian trajectory prediction based on deep convolutional lstm
  network,'' \emph{IEEE Transactions on Intelligent Transportation Systems},
  vol.~22, no.~6, pp. 3285--3302, 2021.

\bibitem{MA20102101}
J.~Ma, W.~Song, J.~Zhang, S.~Lo, and G.~Liao, ``k-nearest-neighbor interaction
  induced self-organized pedestrian counter flow,'' \emph{Physica A:
  Statistical Mechanics and its Applications}, vol. 389, no.~10, pp.
  2101--2117, 2010.

\bibitem{10.1371/journal.pone.0010047}
M.~Moussaïd, N.~Perozo, S.~Garnier, D.~Helbing, and G.~Theraulaz, ``The
  walking behaviour of pedestrian social groups and its impact on crowd
  dynamics,'' \emph{PLOS ONE}, vol.~5, no.~4, pp. 1--7, 04 2010.

\bibitem{BELLOMO20161}
N.~Bellomo, D.~Clarke, L.~Gibelli, P.~Townsend, and B.~Vreugdenhil, ``Human
  behaviours in evacuation crowd dynamics: From modelling to “big data”
  toward crisis management,'' \emph{Physics of Life Reviews}, vol.~18, pp.
  1--21, 2016.

\bibitem{WANG2023128411}
G.~Wang, T.~Chen, H.~Zheng, J.~Wang, X.~Hu, K.~Deng, Z.~Tao, and N.~Luo,
  ``Heterogeneous crowd dynamics considering the impact of personality traits
  under a fire emergency: A questionnaire \& simulation-based approach,''
  \emph{Physica A: Statistical Mechanics and its Applications}, vol. 610, p.
  128411, 2023.

\bibitem{doi:10.1098/rsif.2016.0414}
M.~Moussaïd, M.~Kapadia, T.~Thrash, R.~W. Sumner, M.~Gross, D.~Helbing, and
  C.~Hölscher, ``Crowd behaviour during high-stress evacuations in an
  immersive virtual environment,'' \emph{Journal of The Royal Society
  Interface}, vol.~13, no. 122, p. 20160414, 2016.

\bibitem{SONG2018827}
X.~Song, D.~Han, J.~Sun, and Z.~Zhang, ``A data-driven neural network approach
  to simulate pedestrian movement,'' \emph{Physica A: Statistical Mechanics and
  its Applications}, vol. 509, pp. 827--844, 2018.

\bibitem{ZHAO2020123825}
X.~Zhao, L.~Xia, J.~Zhang, and W.~Song, ``Artificial neural network based
  modeling on unidirectional and bidirectional pedestrian flow at straight
  corridors,'' \emph{Physica A: Statistical Mechanics and its Applications},
  vol. 547, p. 123825, 2020.

\bibitem{8790989}
Y.~Ma, E.~W. Lee, Z.~Hu, M.~Shi, and R.~K. Yuen, ``An intelligence-based
  approach for prediction of microscopic pedestrian walking behavior,''
  \emph{IEEE Transactions on Intelligent Transportation Systems}, vol.~20,
  no.~10, pp. 3964--3980, 2019.

\bibitem{7464842}
Y.~Ma, E.~W.~M. Lee, and R.~K.~K. Yuen, ``An artificial intelligence-based
  approach for simulating pedestrian movement,'' \emph{IEEE Transactions on
  Intelligent Transportation Systems}, vol.~17, no.~11, pp. 3159--3170, 2016.

\bibitem{ZHAO2021103260}
X.~Zhao, J.~Zhang, and W.~Song, ``A radar-nearest-neighbor based data-driven
  approach for crowd simulation,'' \emph{Transportation Research Part C:
  Emerging Technologies}, vol. 129, p. 103260, 2021.

\bibitem{9898931}
H.~Li, Z.~Liu, and B.~Zhou, ``Modeling analysis of t-shaped crowd flow based on
  artificial neural network,'' in \emph{CIBDA 2022; 3rd International
  Conference on Computer Information and Big Data Applications}, 2022, pp.
  1--5.

\bibitem{YAO2020173}
Z.~Yao, G.~Zhang, D.~Lu, and H.~Liu, ``Learning crowd behavior from real data:
  A residual network method for crowd simulation,'' \emph{Neurocomputing}, vol.
  404, pp. 173--185, 2020.

\bibitem{doi:10.1126/sciadv.aay0792}
R.~Bastien and P.~Romanczuk, ``A model of collective behavior based purely on
  vision,'' \emph{Science Advances}, vol.~6, no.~6, p. eaay0792, 2020.

\bibitem{6795963}
S.~Hochreiter and J.~Schmidhuber, ``Long short-term memory,'' \emph{Neural
  Computation}, vol.~9, no.~8, pp. 1735--1780, 1997.

\bibitem{cho-etal-2014-properties}
K.~Cho, B.~van Merri{\"e}nboer, D.~Bahdanau, and Y.~Bengio, ``On the properties
  of neural machine translation: Encoder{--}decoder approaches,'' in
  \emph{Proceedings of {SSST}-8, Eighth Workshop on Syntax, Semantics and
  Structure in Statistical Translation}.\hskip 1em plus 0.5em minus 0.4em\relax
  Doha, Qatar: Association for Computational Linguistics, Oct. 2014, pp.
  103--111.

\bibitem{bai2018empirical}
S.~Bai, J.~Z. Kolter, and V.~Koltun, ``An empirical evaluation of generic
  convolutional and recurrent networks for sequence modeling,'' \emph{arXiv
  preprint arXiv:1803.01271}, 2018.

\bibitem{9671488}
S.~Gopali, F.~Abri, S.~Siami-Namini, and A.~S. Namin, ``A comparison of tcn and
  lstm models in detecting anomalies in time series data,'' in \emph{2021 IEEE
  International Conference on Big Data (Big Data)}, 2021, pp. 2415--2420.

\bibitem{werbos1975beyond}
P.~Werbos, \emph{Beyond Regression: New Tools for Prediction and Analysis in
  the Behavioral Sciences}.\hskip 1em plus 0.5em minus 0.4em\relax Harvard
  University, 1975.

\bibitem{6251245}
M.~Nasir, S.~Nahavandi, and D.~Creighton, ``Fuzzy simulation of pedestrian
  walking path considering local environmental stimuli,'' in \emph{2012 IEEE
  International Conference on Fuzzy Systems}, 2012, pp. 1--6.

\bibitem{XIE2022105875}
W.~Xie, E.~W.~M. Lee, and Y.~Y. Lee, ``Self-organisation phenomena in
  pedestrian counter flows and its modelling,'' \emph{Safety Science}, vol.
  155, p. 105875, 2022.

\bibitem{Yanagisawa_2009}
D.~Yanagisawa, A.~Kimura, A.~Tomoeda, R.~Nishi, Y.~Suma, K.~Ohtsuka, and
  K.~Nishinari, ``Introduction of frictional and turning function for
  pedestrian outflow with an obstacle,'' \emph{Physical Review E}, vol.~80,
  no.~3, sep 2009.

\bibitem{Courtine2003}
G.~Courtine and M.~Schieppati, ``Human walking along a curved path. i. body
  trajectory, segment orientation and the effect of vision,'' \emph{European
  Journal of Neuroscience}, vol.~18, no.~1, pp. 177--190, 2003.

\bibitem{LIU201744}
C.~Liu, W.~Song, L.~Fu, L.~Lian, and S.~Lo, ``Experimental study on relaxation
  time in direction changing movement,'' \emph{Physica A: Statistical Mechanics
  and its Applications}, vol. 468, pp. 44--52, 2017.

\bibitem{MA20102160}
J.~Ma, W.~Song, Z.~Fang, S.~Lo, and G.~Liao, ``Experimental study on
  microscopic moving characteristics of pedestrians in built corridor based on
  digital image processing,'' \emph{Building and Environment}, vol.~45, no.~10,
  pp. 2160--2169, 2010.

\bibitem{Moussa_d_2009}
M.~Moussaïd, D.~Helbing, S.~Garnier, A.~Johansson, M.~Combe, and G.~Theraulaz,
  ``Experimental study of the behavioural mechanisms underlying
  self-organization in human crowds,'' \emph{Proceedings of the Royal Society
  B: Biological Sciences}, vol. 276, no. 1668, pp. 2755--2762, may 2009.

\bibitem{RevModPhys.73.1067}
D.~Helbing, ``Traffic and related self-driven many-particle systems,''
  \emph{Rev. Mod. Phys.}, vol.~73, pp. 1067--1141, Dec 2001.

\bibitem{XIE2022104100}
W.~Xie, E.~W.~M. Lee, and Y.~Y. Lee, ``Simulation of spontaneous
  leader–follower behaviour in crowd evacuation,'' \emph{Automation in
  Construction}, vol. 134, p. 104100, 2022.

\bibitem{drury2009cooperation}
J.~Drury, C.~Cocking, S.~Reicher, A.~Burton, D.~Schofield, A.~Hardwick,
  D.~Graham, and P.~Langston, ``Cooperation versus competition in a mass
  emergency evacuation: A new laboratory simulation and a new theoretical
  model,'' \emph{Behavior research methods}, vol.~41, no.~3, pp. 957--970,
  2009.

\bibitem{PhysRevE.71.036121}
A.~Nakayama, K.~Hasebe, and Y.~b.~u. Sugiyama, ``Instability of pedestrian flow
  and phase structure in a two-dimensional optimal velocity model,''
  \emph{Phys. Rev. E}, vol.~71, p. 036121, Mar 2005.

\bibitem{wkas2006social}
J.~W{\k{a}}s, B.~Gudowski, and P.~J. Matuszyk, ``Social distances model of
  pedestrian dynamics,'' in \emph{Cellular Automata}, S.~El~Yacoubi,
  B.~Chopard, and S.~Bandini, Eds.\hskip 1em plus 0.5em minus 0.4em\relax
  Berlin, Heidelberg: Springer Berlin Heidelberg, 2006, pp. 492--501.

\bibitem{PhysRevEYYF}
Y.~F. Yu and W.~G. Song, ``Cellular automaton simulation of pedestrian counter
  flow considering the surrounding environment,'' \emph{Phys. Rev. E}, vol.~75,
  p. 046112, Apr 2007.

\bibitem{salimans2016weight}
T.~Salimans and D.~P. Kingma, ``Weight normalization: A simple
  reparameterization to accelerate training of deep neural networks,''
  \emph{Advances in neural information processing systems}, vol.~29, 2016.

\bibitem{nair2010rectified}
V.~Nair and G.~E. Hinton, ``Rectified linear units improve restricted boltzmann
  machines,'' in \emph{Icml}, 2010.

\bibitem{srivastava2014dropout}
N.~Srivastava, G.~Hinton, A.~Krizhevsky, I.~Sutskever, and R.~Salakhutdinov,
  ``Dropout: a simple way to prevent neural networks from overfitting,''
  \emph{The journal of machine learning research}, vol.~15, no.~1, pp.
  1929--1958, 2014.

\bibitem{Cao_2017}
S.~Cao, A.~Seyfried, J.~Zhang, S.~Holl, and W.~Song, ``Fundamental diagrams for
  multidirectional pedestrian flows,'' \emph{Journal of Statistical Mechanics:
  Theory and Experiment}, vol. 2017, no.~3, p. 033404, mar 2017.

\bibitem{su11195501}
C.~Dias, M.~Abdullah, M.~Sarvi, R.~Lovreglio, and W.~Alhajyaseen, ``Modeling
  and simulation of pedestrian movement planning around corners,''
  \emph{Sustainability}, vol.~11, no.~19, 2019.

\bibitem{Ye_2019}
R.~Ye, M.~Chraibi, C.~Liu, L.~Lian, Y.~Zeng, J.~Zhang, and W.~Song,
  ``Experimental study of pedestrian flow through right-angled corridor: uni-
  and bidirectional scenarios,'' \emph{Journal of Statistical Mechanics: Theory
  and Experiment}, vol. 2019, no.~4, p. 043401, apr 2019.

\bibitem{Zhang_2011}
J.~Zhang, W.~Klingsch, A.~Schadschneider, and A.~Seyfried, ``Transitions in
  pedestrian fundamental diagrams of straight corridors and t-junctions,''
  \emph{Journal of Statistical Mechanics: Theory and Experiment}, vol. 2011,
  no.~06, p. P06004, jun 2011.

\bibitem{LI2022126593}
Y.~Li, X.~Yang, Z.~Wang, L.~Chen, and Y.~Chen, ``Lane-design for mixed
  pedestrian flow in t-shaped passage,'' \emph{Physica A: Statistical Mechanics
  and its Applications}, vol. 589, p. 126593, 2022.

\bibitem{Cao_2018}
S.~Cao, J.~Zhang, W.~Song, C.~Shi, and R.~Zhang, ``The stepping behavior
  analysis of pedestrians from different age groups via a single-file
  experiment,'' \emph{Journal of Statistical Mechanics: Theory and Experiment},
  vol. 2018, no.~3, p. 033402, mar 2018.

\bibitem{Savitzky_smooth}
A.~Savitzky and M.~J.~E. Golay, ``Smoothing and differentiation of data by
  simplified least squares procedures.'' \emph{Analytical Chemistry}, vol.~36,
  no.~8, pp. 1627--1639, 1964.

\bibitem{7780479}
A.~Alahi, K.~Goel, V.~Ramanathan, A.~Robicquet, L.~Fei-Fei, and S.~Savarese,
  ``Social lstm: Human trajectory prediction in crowded spaces,'' in \emph{2016
  IEEE Conference on Computer Vision and Pattern Recognition (CVPR)}, 2016, pp.
  961--971.

\bibitem{DBLP}
A.~Vemula, K.~M{\"{u}}lling, and J.~Oh, ``Social attention: Modeling attention
  in human crowds,'' \emph{CoRR}, vol. abs/1710.04689, 2017.

\bibitem{doi:10.1086/200975}
E.~T. Hall, R.~L. Birdwhistell, B.~Bock, P.~Bohannan, A.~R. Diebold, M.~Durbin,
  M.~S. Edmonson, J.~L. Fischer, D.~Hymes, S.~T. Kimball, W.~La~Barre, , J.~E.
  McClellan, D.~S. Marshall, G.~B. Milner, H.~B. Sarles, G.~L. Trager, and
  A.~P. Vayda, ``Proxemics [and comments and replies],'' \emph{Current
  Anthropology}, vol.~9, no. 2/3, pp. 83--108, 1968.

\bibitem{kingma2017adam}
D.~P. Kingma and J.~Ba, ``Adam: A method for stochastic optimization,'' 2017.

\bibitem{STEFFEN20101902}
B.~Steffen and A.~Seyfried, ``Methods for measuring pedestrian density, flow,
  speed and direction with minimal scatter,'' \emph{Physica A: Statistical
  Mechanics and its Applications}, vol. 389, no.~9, pp. 1902--1910, 2010.

\bibitem{08bba04d17ad491abd48adb734a4fadd}
H.~Xue, D.~Huynh, and M.~Reynolds, ``\BIBforeignlanguage{English}{Ss-lstm: A
  hierarchical lstm model for pedestrian trajectory prediction},'' in
  \emph{\BIBforeignlanguage{English}{2018 IEEE Winter Conference on
  Applications of Computer Vision (WACV)}}.\hskip 1em plus 0.5em minus
  0.4em\relax United States: IEEE, Institute of Electrical and Electronics
  Engineers, May 2018, pp. 1186--1194, 2018 IEEE Winter Conference on
  Applications of Computer Vision, WACV 2018 ; Conference date: 12-03-2018
  Through 15-03-2018.

\bibitem{Social-GAN}
A.~Gupta, J.~Johnson, L.~Fei{-}Fei, S.~Savarese, and A.~Alahi, ``Social {GAN:}
  socially acceptable trajectories with generative adversarial networks,''
  \emph{CoRR}, vol. abs/1803.10892, 2018.

\bibitem{Scene-LSTM}
H.~Manh and G.~Alaghband, ``Scene-lstm: {A} model for human trajectory
  prediction,'' \emph{CoRR}, vol. abs/1808.04018, 2018.

\end{thebibliography}

\begin{IEEEbiography}[{\includegraphics[width=1in,height=1.25in,clip,keepaspectratio]{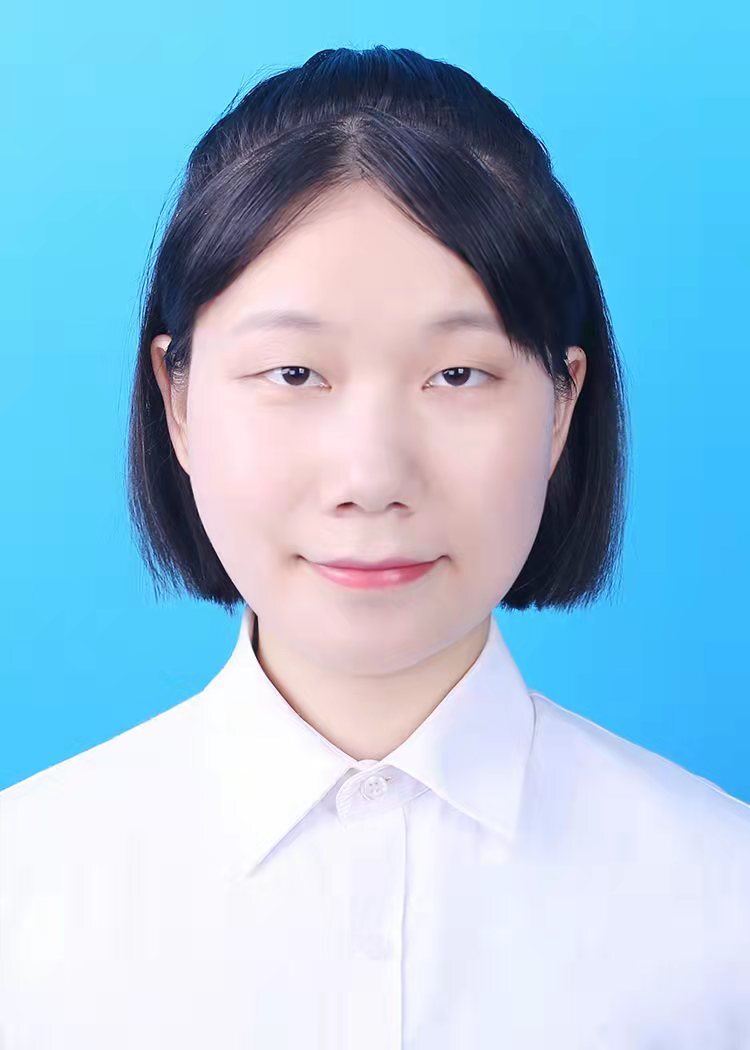}}]{Xuanwen Liang} 
\quad received the bachelor’s degree from Southwest Jiaotong University, and the master’s degree from University of Science and Technology of China. She is currently pursuing the Ph.D. degree with the Department of Architecture and Civil Engineering, City University of Hong Kong. Her research interests include data-driven crowd simulation, deep learning, pedestrian and evacuation dynamics.
\end{IEEEbiography}

\begin{IEEEbiography}[{\includegraphics[width=1in,height=1.25in,clip,keepaspectratio]{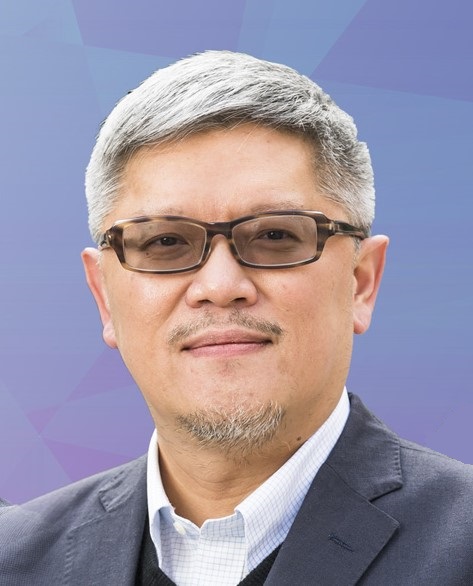}}]{Eric Wai Ming Lee} 
\quad received the B.Eng. degree in building services engineering and the Ph.D. degree in fire engineering from the City University of Hong Kong. He is currently a Professor with the Department of Architecture and Civil Engineering, City University of Hong Kong. His research interests include building evacuation, pedestrian movement, and human behavior in movement. He is a professional engineer and currently the panel member of Fire Discipline of the Hong Kong Institution of Engineers, the assessor of Hong Kong Laboratory Accreditation Scheme of the HKSAR Government for fire testing laboratories and also the member of various fire safety related technical committees of the Buildings Department and Fire Services Department of the HKSAR Government.

\end{IEEEbiography}

\end{document}